\documentclass{article}

    \usepackage[final,nonatbib]{neurips_data_2021}

\usepackage[utf8]{inputenc} 
\usepackage[T1]{fontenc}    
\usepackage{url}            
\usepackage{booktabs}       
\usepackage{amsfonts}       
\usepackage{nicefrac}       
\usepackage{microtype}      
\usepackage{xcolor}         

\usepackage[pdftex]{graphicx}
\usepackage{multirow}
\usepackage{booktabs}
\usepackage{verbatim}
\usepackage{subfig}
\usepackage{bm}
\usepackage{adjustbox}
\usepackage{footnote}
\makesavenoteenv{tabular}
\makesavenoteenv{table}

\usepackage[font=small,labelfont=bf]{caption}  
\usepackage{makecell}
\usepackage{tabulary}
\definecolor{demphcolor}{RGB}{144,144,144}

\definecolor{mygray}{gray}{0.4}
\usepackage{pifont}
\newlength\savewidth\newcommand\shline{\noalign{\global\savewidth\arrayrulewidth
  \global\arrayrulewidth 1pt}\hline\noalign{\global\arrayrulewidth\savewidth}}

\newcommand{\tablestyle}[2]{\setlength{\tabcolsep}{#1}\renewcommand{\arraystretch}{#2}\centering\footnotesize}
\makeatletter\renewcommand\paragraph{\@startsection{paragraph}{4}{\z@}
  {.5em \@plus1ex \@minus.2ex}{-.5em}{\normalfont\normalsize\bfseries}}\makeatother

\newcolumntype{C}[1]{>{\centering\arraybackslash}p{#1}}
\newcolumntype{R}[1]{>{\raggedleft\arraybackslash}p{#1}}
\newcolumntype{L}[1]{>{\raggedright\arraybackslash}p{#1}}

\newcommand{\newcellc}[2][c]{%
  \begin{tabular}[#1]{@{}c@{}}#2\end{tabular}}
\newcommand{\newcelll}[2][l]{%
  \begin{tabular}[#1]{@{}l@{}}#2\end{tabular}}
  \newcommand{\newcellr}[2][r]{%
  \begin{tabular}[#1]{@{}r@{}}#2\end{tabular}}

\definecolor{bittersweet}{rgb}{1.0, 0.44, 0.37}

\definecolor{mygreen}{rgb}{0.29, 0.7, 0.48}

\definecolor{mbccolor}{rgb}{0.59, 0.1, 0.18}

\makeatletter
\newcommand\footnoteref[1]{\protected@xdef\@thefnmark{\ref{#1}}\@footnotemark}
\makeatother

\usepackage{tikz}
\newcommand{\Arrow}[1]{%
\parbox{#1}{\tikz{\draw[->](0,0)--(#1,0);}}
}
\usepackage{arydshln}

\definecolor{citecolor}{RGB}{34,139,34} 
\usepackage[pagebackref=true,breaklinks=true,colorlinks,
citecolor=citecolor,bookmarks=false]{hyperref}

\title{VALUE: A Multi-Task Benchmark for Video-and-Language Understanding Evaluation}

\author{%

 Linjie Li\footnotemark[1]\,\,$^1$, Jie Lei\thanks{\, Equal contribution.}\,\,$^2$, Zhe Gan$^1$, Licheng Yu$^2$, Yen-Chun Chen$^1$, \\ \bf  Rohit Pillai$^1$, Yu Cheng$^1$, Luowei Zhou$^1$,  Xin Eric Wang$^3$, William Yang Wang$^4$, \\ \bf Tamara L. Berg$^2$, Mohit Bansal$^2$, Jingjing Liu$^5$, Lijuan Wang$^1$, Zicheng Liu$^1$\\ 
$^1$Microsoft Corporation\quad $^2$UNC Chapel Hill \\
$^3$UC Santa Cruz \quad $^4$UC Santa Barbara \quad $^5$Tsinghua University\\
\texttt{\small\{lindsey.li,zhe.gan,yen-chun.chen,rohit.pillai,}\\ 
\texttt{\small yu.cheng,luowei.zhou,lijuanw,zliu\}@microsoft.com} \\
\texttt{\small\{jielei,licheng,tlberg,mbansal\}@cs.unc.edu} \\
\texttt{\small xwang366@ucsc.edu, william@cs.ucsb.edu, JJLiu@air.tsinghua.edu.cn}\\
}

\begin{document}

\maketitle

\begin{abstract}
Most existing video-and-language (VidL) research focuses on a single dataset, or multiple datasets of a single task. In reality, a truly useful VidL system is expected to be easily generalizable to diverse tasks, domains, and datasets.
To facilitate the evaluation of such systems, we introduce \textbf{V}ideo-\textbf{A}nd-\textbf{L}anguage \textbf{U}nderstanding \textbf{E}valuation (\textbf{VALUE}) benchmark, an assemblage of 
11 VidL
datasets over 3 popular tasks: ($i$) text-to-video retrieval; ($ii$) video question answering; and ($iii$) video captioning. 
VALUE benchmark aims to cover a broad range of video genres, video lengths, data volumes, and task difficulty levels. 
Rather than focusing on single-channel videos with visual information only, VALUE promotes models that leverage information from both video frames and their associated subtitles, as well as models that share knowledge across multiple tasks.
We evaluate various baseline methods with and without large-scale VidL pre-training, and systematically investigate the impact of video input channels, fusion methods, and different video representations. We also study the transferability between tasks and conduct multi-task learning under different settings. 
The significant gap between our best model and human performance calls for future study for advanced VidL models. 
VALUE is available at \url{https://value-benchmark.github.io/}.\footnote{VALUE competition will be held in conjunction with CLVL workshop at ICCV 2021, \url{https://sites.google.com/view/iccv21clvl/home}.
}
\end{abstract}

\vspace{-1mm}
\section{Introduction}
\vspace{-1mm}
Joint video-and-language (VidL) understanding sits at the nexus of computer vision and natural language processing (NLP), and has attracted rapidly growing attention from both communities. Popular tasks include  
text-based video retrieval~\cite{xu2016msr, krishna2017dense, rohrbach2015dataset,lei2020tvr,li2020hero}, video moment retrieval~\cite{anne2017localizing, gao2017tall, krishna2017dense, lei2020tvr,li2020hero}, video question answering~\cite{xu2017video, jang2017tgif, lei2018tvqa,lei2019tvqa+}, and video captioning~\cite{rohrbach2015dataset,xu2016msr,zhou2018towards,lei2020tvr}. 
However, existing works~\cite{liu2019use,jiang2020divide,le2020hierarchical,gabeur2020multi,patrick2020support,luo2021clip4clip} in this field are often evaluated on distinct datasets under different experimental settings, making fair comparison difficult between methods.
Meanwhile, most works are evaluated on a limited set of tasks, thus difficult to measure as a universal VidL system.
As exemplary, in the NLP community, GLUE~\cite{wang2018glue} and SuperGLUE~\cite{wang2019superglue} have evolved into prominent evaluation frameworks that continue to push the frontier of natural language understanding, due to their broad coverage of NLP tasks with diverse training data volumes, task genres, and unified task formulation.

Inspired by them, to better benchmark advances in VidL research, we introduce \textbf{V}ideo-\textbf{A}nd-\textbf{L}anguage \textbf{U}nderstanding \textbf{E}valuation (\textbf{VALUE}) benchmark, an online platform with a compilation of 11 VidL datasets for model evaluation and comparison. There are several contributions that render VALUE a unique and valuable asset to the community. ($i$) \textbf{Diversity}: To evaluate the versatility and generalizability of VidL systems, our benchmark includes diverse tasks, including video retrieval, question answering (QA), and captioning (see Section~\ref{sec:dataset_and_evaluation} for details). VALUE also covers a broad range of video genres, video lengths, and data volumes. 
($ii$) \textbf{Multi-channel video inputs}: Videos are multi-channeled and usually contain frames, audio, and textual information. Most of the existing works, however, only focus on the use of video frames. In our benchmark, we provide both video frames and their accompanying dialogues in the form of subtitle sentences\footnote{ASR can be applied when subtitles are not available.} 
as video inputs. Tasks that require multi-channel information for inference are preferable. In TVQA~\cite{lei2018tvqa}, for example, the cues to answering the questions are usually in both visual and dialogue content.
($iii$) \textbf{Task difficulties}: Our benchmark is challenging and hard-to-game. We found that even the best VidL models we tested underperform human baselines by a large margin, suggesting great space for improvement. ($iv$) \textbf{Easy evaluation}: For each dataset, we select a representative metric from a set of standard metrics for evaluation. We divide the datasets into 3 categories, and rank participants in each category based on the meta-average score across associated tasks. For the VALUE leaderboard, we provide a universal target metric (\emph{i.e.}, the meta-average score across all the tasks) to track progress. 
We also release rich pre-extracted video frame features, offer starter code, and withhold private test data for reliable evaluation on our evaluation server. 

To provide an in-depth analysis of our VALUE benchmark, we evaluate a number of baselines with and without pre-training, and systematically assess the effects of video input channels, fusion methods, and different video representations. 
We also investigate the transferability between tasks and the effect of multi-task training under various settings (\emph{e.g.}, multi-task learning by task type or by data domain). 
Video-and-language understanding is challenging, as it encompasses a wide range of areas such as visual and linguistic semantic understanding, spatial-temporal grounding, multimodal fusion, and commonsense reasoning.
We envision that VALUE will inspire active research and discussion in the community. More details are available at \url{https://value-benchmark.github.io/}.

\begin{table}[!t]
\caption{
\textbf{Statistics of video data} used in VALUE benchmark. Multi-channel ratio refers to percentage of videos with subtitles. Video lengths are measured in terms of seconds (s) on average.
}
\vspace{2mm}

\tablestyle{5pt}{1.1} 
\def\w{30pt} 
\scalebox{1.0}{
\begin{tabular}{l|rrrr}
Video Data & Source & \#Video & Multi-channel Ratio  & Length \\
\shline
TV (TVQA, TVR, TVC) & TV episodes & 21.8K & 100\% & 76s \\
How2 (How2R, How2QA) & Instructional Videos on Youtube & 31.7K & 99.36\% & 59s \\
VIOLIN & TV episodes, Movie Clips & 15.9K & 99.33\% & 40s \\
VLEP & TV episodes, Vlog on Youtube & 10.2K & 98.11\% & 32s \\
YouCook2 (YC2C, YC2R) & Cooking Videos on Youtube & 15.4K & 94.40\% & 20s \\
VATEX (VATEX-EN-R/C) & Various Youtube Videos & 41.3K & 50.93\% & 10s \\
\end{tabular}
}
\vspace{-2mm}
\label{tab:video_stats}
\end{table}

\vspace{-2mm}
\section{Related Work}
\vspace{-1mm}
Publicly accessible large-scale multi-task benchmarks~\cite{conneau2018senteval,wang2018glue,wang2019superglue,hu2020xtreme,liang2020xglue} have facilitated recent advances~\cite{devlin2018bert,yang2019xlnet,liu2019roberta,xue2020mt5,wei2020learning,fang2020filter,ouyang2020ernie} in NLP. For example,
SentEval~\cite{conneau2018senteval} contains a collection of natural language tasks, such as sentiment analysis~\cite{pang2005seeing,socher2013recursive}, entailment~\cite{bowman2015large} and semantic textual similarity~\cite{agirre2012semeval,agirre2013sem,agirre2014semeval,agirre2015semeval,agirre2016semeval}.
While SentEval aims at evaluating sentence-level vector representations, GLUE~\cite{wang2018glue} advanced it by removing all restrictions on the model -- GLUE is designed to be model-agnostic, allowing the evaluation of any type of representation. 
With the introduction of large-scale transformer~\cite{vaswani2017attention} language models such as BERT~\cite{devlin2018bert}, RoBERTa~\cite{liu2019roberta}, XLNET~\cite{yang2019xlnet} and OpenAI GPT~\cite{radford2018improving}, the headroom of GLUE is drastically decreasing.
SuperGLUE~\cite{wang2019superglue} was later proposed as a more rigorous test for language understanding, which incorporates more challenging and diverse tasks. XTREME~\cite{hu2020xtreme} and XGLUE~\cite{liang2020xglue} have also been proposed for benchmarking multilingual language understanding. 
Our VALUE benchmark shares similar merit to these language understanding benchmarks, focusing on understanding and generation tasks in the video-and-language domain.

Compared to the blossoming of natural language benchmarks,  video-and-language (VidL) understanding still lacks a large-scale benchmark to systematically track advances in this area.
Methods developed~\cite{sun2019videobert,zhu2020actbert,miech2020end,liu2019use,jiang2020divide,le2020hierarchical,li2020hero,gabeur2020multi,patrick2020support,lei2021less,tang2021decembert,luo2021clip4clip,lin2021vx2text,zhou2021cupid} in this field are often evaluated on different tasks, datasets and experimental settings, making fair comparison difficult.
The Pentathlon Challenge~\cite{albanie2020end} held at CVPR 2020 combines 5 text-based video retrieval tasks to compare models using a set of pre-extracted expert features.
However, the Challenge only focuses on a single task, and limits the models to only using offline extracted features.
In contrast, VALUE is designed to incorporate a diverse set of tasks, including text-based video retrieval~\cite{zhou2018towards,wang2019vatex}, video moment retrieval~\cite{lei2020tvr,li2020hero}, video question answering~\cite{lei2018tvqa,li2020hero}, video captioning~\cite{zhou2018towards,wang2019vatex,lei2020tvr}, video-and-language inference~\cite{liu2020violin}, and next event prediction~\cite{lei2020more}.
Meanwhile, VALUE is model-agnostic and welcomes methods of all kinds.

\begin{table}[!t]
\caption{
\textbf{Statistics of datasets} in VALUE benchmark. Ground-truth annotations on Test (leaderboard) split are hidden from the public, and used to rank model performance. ($\dag$) VIOLIN and VLEP are 2-way classification tasks, which are considered as special QA tasks in our benchmark for simplicity. 
\textit{AveR} denotes Average Recall at \{1, 5, 10\}, \textit{Acc.} = Classification Accuracy.
}
\vspace{2mm}
\tablestyle{5pt}{1.1} 
\def\w{30pt} 
\resizebox{\textwidth}{!}{
\begin{tabular}{l|l|cccc|L{\w}}
Task  &  Dataset  & \multicolumn{4}{c|}{Data Statistics (\# videos/ \# queries, QAs, captions)} &  Metrics  \\
\cline{3-6}
& & Train & Val & Test (public) & Test (leaderboard)\footnote{The ground-truth annotations on Test (leaderboard) split are either obtained from the author or collected following the same procedure as in the original paper.} & \\
\shline
\multirow{4}{*}{ Retrieval } & TVR~\cite{lei2020tvr} & 17.4K/87.1K & 2.2K/10.9K & - & 2.2K/10.9K & \multirow{4}{*}{ AveR } \\
& How2R~\cite{li2020hero} & 21.3K/27.1K & 1.0K/1.3K & - & 1.0K/1.3K & \\
& YC2R~\cite{zhou2018towards} & 10.3K/10.3K & 3.5K/3.5K & - & 1.6K/1.6K & \\
& VATEX-EN-R~\cite{wang2019vatex} & 26.0K/259.9K & 3.0K/30.0K & - & 6.0K/60.0K & \\
\hline
\multirow{4}{*}{ QA } & TVQA~\cite{lei2018tvqa} & 17.4K/122.0K & 2.2K/15.3K & - & 2.2K/15.3K & \multirow{4}{*}{ Acc. } \\
& How2QA~\cite{li2020hero} & 24.5K/34.2K & 3.1K/3.1K & - & 3.1K/3.1K & \\
\cdashline{2-6}
& VIOLIN$^\dag$~\cite{liu2020violin} & 12.7K/76.1K & 1.6K/9.6K & 1.6K/9.6K & 1.3K/7.7K & \\
& VLEP$^\dag$~\cite{lei2020more} & 7.2K/20.1K & 1.6K/4.4K & - & 1.5K/4.2K & \\
\hline
\multirow{3}{*}{ Captioning } & TVC~\cite{lei2020tvr} & 17.4K/86.7K & 10.8K/43.6K & - & 10.8K/43.6K & \multirow{3}{*}{\newcelll{CIDEr-D}} \\
& YC2C~\cite{zhou2018towards} & 10.3K/10.3K & 3.5K/3.5K & - & 1.6K/1.6K & \\
& VATEX-EN-C~\cite{wang2019vatex} & 26.0K/259.9K & 3.0K/30.0K & 6.0K/60.0K  & 6.2K/62.8K  & \\
\end{tabular}
}
\vspace{-3mm}
\label{tab:data_stats}
\end{table}

\vspace{-2mm}
\section{VALUE Benchmark Tasks}\label{sec:dataset_and_evaluation}
\vspace{-1mm}
VALUE aims to provide a one-stop evaluation for multi-channel video understanding on 3 common video-and-language (VidL) tasks: ($i$) text-based video retrieval; ($ii$) video question answering (QA); and ($iii$) video captioning. To construct a comprehensive evaluation benchmark, we include recent datasets collected on multi-channel videos: TVR~\cite{lei2020tvr}, How2R~\cite{li2020hero}, TVQA~\cite{lei2018tvqa}, How2QA~\cite{li2020hero}, VIOLIN~\cite{liu2020violin}, VLEP~\cite{lei2020more} and TVC~\cite{lei2020tvr}. Since most of these datasets focus on understanding long videos in TV/movie domain, we further select another two popular datasets, YouCook2~\cite{zhou2018towards} and VATEX~\cite{wang2019vatex}, originally built on shorter single-channel YouTube videos, to cover diverse video genres and lengths. 
In total, VALUE assembles 11 diverse VidL datasets.\footnote{Video features, subtitles and annotations for all the VALUE tasks are released at \url{https://github.com/value-benchmark/DataRelease}. Due to copyright issue, we are unable to publish raw videos. However, we provide all the YouTube ids/TV episode versions along with their original timestamps to facilitate end-to-end training on VALUE benchmark.}
There are other VidL datasets on single-channel videos that are not included in VALUE, due to the difficulties in collecting hidden test set~\cite{msvd,xu2016msr,didemo}, unnatural annotations~\cite{msrvtt-qa},
or the lack of subtitle channel in GIF videos~\cite{jang2017tgif}. 

\begin{figure}[h]
    \centering
    \includegraphics[width=\textwidth]{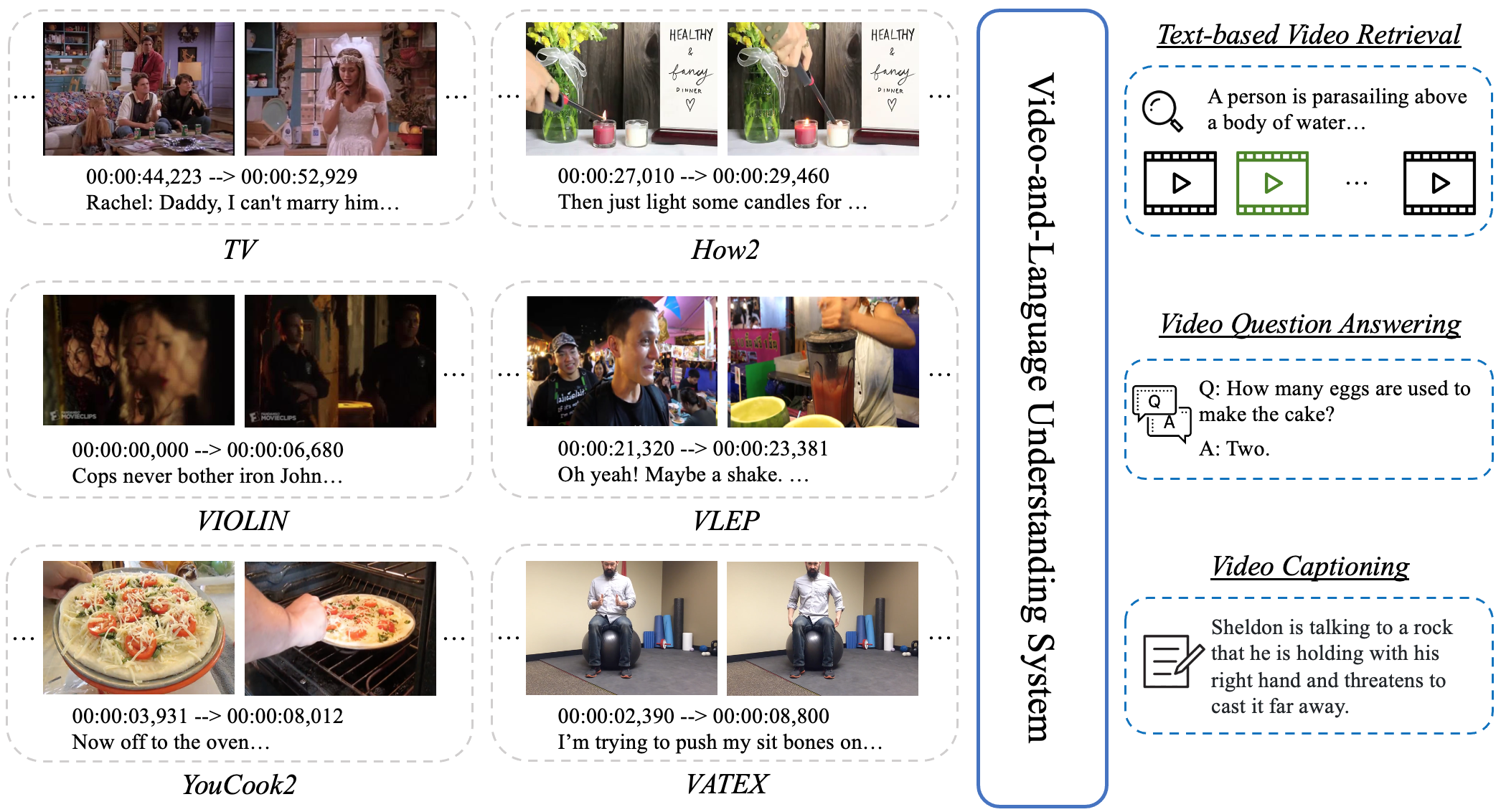}
    \caption{An illustration of VALUE benchmark. VALUE provides a one-stop evaluation for multi-channel video understanding on 3 common video-and-language tasks: ($i$) text-based video retrieval; ($ii$) video question answering; and ($iii$) video captioning. The videos in VALUE come from 6 data sources, covering diverse genres and domains.}
    \label{fig:value}
\end{figure}

Table~\ref{tab:video_stats} summarizes the statistics of video data provided in VALUE. The videos come from diverse domains, ranging from TV episodes and movie clips with different temporal dynamics, event shifts and people interactions, to instructional videos and vlogs dominated by monologues with less human-centered scenes. Average video length varies from 10 to 76 seconds. All the video datasets except VATEX~\cite{wang2019vatex} have a high multi-channel ratio (proportion of videos to subtitles). Table~\ref{tab:data_stats} summarizes the selected tasks and datasets. Figure~\ref{fig:value} shows an illustration of the VALUE benchmark.

Our VALUE evaluation server is hosted on CodaLab.\footnote{See our submission page for details: \url{https://value-benchmark.github.io/submission.html}.} 
In the following subsections, we will introduce each task.
The benchmark site shows the scores per-task and a meta-average of those scores across all tasks to determine a system’s rank on the leaderboard.

\subsection{Text-based Video Retrieval Tasks} 
In VALUE, there are two types of text-based video retrieval tasks: ($i$) Video Corpus Moment Retrieval (VCMR): TVR and How2R datasets; and ($ii$) Video Retrieval (VR): YouCook2 Retrieval (YC2R) and VATEX Retrieval (VATEX-EN-R) datasets. VR requires a model to retrieve the most relevant video clip from the video corpus described by the textual query. VCMR is more challenging, requiring a model to not only retrieve the most relevant video clip from the video corpus, but also locate the relevant moment in the retrieved video clip. Stand-alone evaluation on temporal moment localization tasks~\cite{anne2017localizing,gao2017tall} is not included in our benchmark, as the two VCMR tasks already evaluate the ability of the model to localize relevant moment as a sub-task. The upper block of Table~\ref{tab:data_stats} summarizes the statistics of the 4 datasets for retrieval tasks.

\textbf{TVR}~\cite{lei2020tvr} consists of 109K queries on 21.8K videos from 6 TV shows of diverse genres, where each query is associated with a tight temporal alignment. 
Among all queries, 74.2\% are related to video only, 9.1\% to text only, and 16.6\% to both video and text. The dataset is divided into 80\% train, 10\% val, 5\% test-public, and 5\% test-private. We combine the test-public set with the test-private set for leaderboard evaluation.

\textbf{How2R}~\cite{li2020hero} is collected following the same procedure of TVR, but based on 60-second clips from 9K instructional videos in HowTo100M~\cite{miech2019howto100m}, on average 2-3 queries per clip. The original How2R data are noisy due to short and repetitive textual queries. For VALUE benchmark, we 
remove queries with fewer than 6 words and repetitions. After cleaning, 2K video clips and the associated queries are held-out for validation and testing, and the rest for training. 

\textbf{YouCook2 Retrieval (YC2R)}~\cite{zhou2018towards} consists of 2K YouTube cooking videos across 89 recipe types. The videos are split into a 67\%/23\%/10\% for
training/validation/testing 
and segmented by human annotators into clips that represent recipe steps. Each clip is annotated with one textual description. We follow~\cite{miech2019howto100m} to evaluate retrieval performance at clip level.
We obtain private labels from the authors to provide evaluation on test set, and augment the videos with ASR-generated subtitles.

\textbf{VATEX Retrieval (VATEX-EN-R)}  VATEX~\cite{wang2019vatex} was originally developed for multilingual video captioning and video-guided machine translation tasks. It contains 41.3K videos of 600 fine-grained human activities and 825K captions in both English and Chinese. To ensure its consistency with other tasks being considered, we take videos and English captions to evaluate retrieval performance. Videos are split into 26K/3K/6K/6K for training/validation/public testing/private testing. We use the public test set to evaluate retrieval performance, to avoid potential data leaks on private testing set of the original VATEX tasks. We also
provide ASR-generated subtitles. 

To evaluate the model performance, we adopt the average recall at K (R@K) over all queries as the metric. 
For VR (\emph{i.e.}, YC2R and VATEX-EN-R), we consider a prediction correct if the predicted video matches the ground-truth video.
For VCMR (\emph{i.e.}, TVR and How2R),
we additionally require that the predicted span of a correct prediction has a high overlap with the ground-truth moment.
We use temporal Intersection over Union (tIoU) to measure the overlap between the predicted span and the ground-truth span.\footnote{\label{note:tvr_eval}During evaluation, the average recalls are measured by tIoU$\geq$0.7. } We use AveR (the average of R@\{1, 5, 10\}) as the final metric to evaluate model performance on retrieval tasks. 

\subsection{Video Question Answering Tasks}
We group tasks that use classification accuracy as the evaluation metric into video question answering (QA) tasks. The middle block of Table~\ref{tab:data_stats} summarizes the data statistics of 4 datasets considered. 

\textbf{TVQA}~\cite{lei2018tvqa} is collected under multiple-choice settings from TV videos.
Each video clip contains 7 questions, with 5 answers per question.  The start/end points of the relevant moments are also provided for each question. TVQA consists of 3 sub-tasks: ($i$) QA on the grounded clip; ($ii$) question-guided moment localization; and ($iii$) QA on the full video clip. We only consider QA on the full video clip, as this is the most challenging setting among the three. We combine the test-public set with the test-private set for leaderboard evaluation.\footnote{\label{note:tv_split}Train, val and test video splits are the same as TVR.}

\textbf{How2QA}~\cite{li2020hero} was collected in a similar way to TVQA, but on video clips sampled from instructional videos in HowTo100M~\cite{miech2019howto100m}. Each video clip is annotated with an average of 1-2 questions, with 4 answers per question. Similarly, the questions in How2QA are grounded temporally, but  
we only consider QA on the full video clip. 
As the video clips used in How2QA largely overlap with those in How2R, 
we re-split the video clips and their associated QA pairs into 80\% train, 10\% val and 10\% test, to avoid potential data leaks. 

\textbf{VIOLIN}~\cite{liu2020violin} is introduced as a new video-and-language inference task. Given a premise video clip with aligned subtitles and a hypothesis sentence, the task is to predict whether the premise entails the hypothesis or contradicts the hypothesis. Its original release consists of 95.3K video-hypothesis pairs with ground-truth annotations from 15.9K video clips, split into 80\% train, 10\% val and 10\% test. We further collect a hidden test split (\emph{i.e.}, Test (leaderboard) in Table~\ref{tab:data_stats}) with 4K hypothesis on 1.5K video clips from the same video domain for leaderboard evaluation.

\textbf{VLEP}~\cite{lei2020more} is a dataset for video-and-language commonsense-based future event prediction. 
Given a video with aligned subtitles, the task is to choose which of the two future events is more likely to occur after that video.
VLEP contains 28.7K future event prediction examples from 10.2K TV shows and YouTube Lifestyle Vlog video clips, which are
split into 70\% train, 15\% val and 15\% test. 

\vspace{-1mm}
\subsection{Video Captioning Tasks} 
\vspace{-1mm}
For video captioning tasks, we consider 3 datasets (lower block of Table~\ref{tab:data_stats}). 

\textbf{TVC}~\cite{lei2020tvr} is a multi-channel video captioning dataset extended from TVR, containing 262K descriptions paired with 108K video moments. 
Unlike traditional video captioning tasks, the descriptions are collected on video moments instead of the entire video, and video subtitles are used as additional model input. 
For a given video and the start/end points of a moment of the video, a model must generate a description for the video moment with/without leveraging the information from the entire video.
We combine the test-public set with the test-private set for leaderboard evaluation.

\textbf{YouCook2 Captioning (YC2C)}~\cite{zhou2018towards} is built on the same cooking videos as in YouCook2 Retrieval task. Each video clip is annotated with one captioning sentence. 
Depending on whether we regard each clip individually or combine the clip captions into a paragraph, the evaluation for each video can be either at clip-level, by reporting the averaged score on each clip over the entire video corpus; or at video-level, evaluating each merged paragraph. 
We follow~\cite{zhou2018end} to evaluate clip-level performance, and to maintain consistency with other captioning datasets considered. The test set is used for leaderboard evaluation.

\textbf{VATEX Captioning (VATEX-EN-C)}~\cite{wang2019vatex}  Similar to VATEX Retrieval, we take videos and English captions in VATEX as another task to evaluate video captioning on multi-channel videos. Each video is annotated with 10 English captions, with 5 regular English captions and 5 parallel English captions translated from Chinese. The private test set is used for leaderboard evaluation. 

The performance of video captioning tasks is measured by comparing predicted captions against corresponding ground-truth captions, with standard captioning metrics applied (\emph{e.g.}, BLEU@4~\cite{papineni2002bleu}, METEOR~\cite{denkowski2014meteor}, ROUGE-L~\cite{linrouge}, and CIDEr-D~\cite{vedantam2015cider}).  We use the CIDEr-D score as the main metric to evaluate model performance.

\vspace{-1mm}
\section{Experiments and Analysis}
\vspace{-1mm}
In this section, we provide extensive experiments and analysis to demonstrate the value of VALUE benchmark. 
Specifically, we investigate the impact of input channels and video-subtitle fusion methods (Sec.~\ref{sec:modality}), evaluate the effectiveness of different visual representations (Sec.~\ref{sec:vis_feat}), and study the transferability between tasks (Sec.~\ref{sec:transfer}) and the impact of multi-task learning (Sec.~\ref{sec:multi-task}). 

\vspace{-1mm}
\subsection{Features and Baseline}

\textbf{Multi-Channel Video Representations.} The context inputs are multi-channel videos, \emph{i.e.}, videos and their associated subtitles. For \textit{subtitle channel}, we follow \cite{liu2019roberta} and tokenize each subtitle sentence 
into a sequence of WordPieces \cite{wu2016google}.
The final representation for each sub-word token is the summation of its token embedding and position embedding, followed by a layer normalization (LN) layer.
For \textit{video channel}, we extract 2D appearance features and 3D motion features every 1.5 seconds.
We use ResNet-152~\cite{he2016deep} pre-trained on ImageNet~\cite{deng2009imagenet} to extract 2D features, and use SlowFast~\cite{feichtenhofer2019slowfast} pre-trained on Kinetics~\cite{kay2017kinetics} to extract 3D features.
These features are concatenated and fed through a fully-connected (FC) layer to be projected into the same lower-dimensional space as token embeddings. 
Since video frames are sequential, their position embeddings can be calculated in the same way as word tokens. 
The final embedding of a video segment is obtained by summing up FC outputs and position embeddings, then passing through an LN layer. 

\textbf{Baseline Architecture.}  There are many pioneering works on building generalizable VidL understanding systems via large-scale pre-training~\cite{miech2019howto100m,zhu2020actbert,miech2020end,sun2019videobert,li2020hero}. However, most focus on single-channel videos, thereby cannot be evaluated directly on or easily extended to multi-channel videos. Our selected baseline architecture is based on HERO~\cite{li2020hero}, due to its strong capacity of understanding multi-channel videos and its generalizability to different VidL tasks.\footnote{Code is released at \url{https://github.com/value-benchmark/StarterCode}.}

HERO takes as inputs a sequence of video segments and subtitle sentences, and encodes them in a hierarchical fashion, with a cross-modal transformer to fuse subtitle sentences and their accompanying local video segments. The cross-modal transformer is followed by a temporal transformer to obtain a globally contextualized embedding for each segment, using all the segments in the video. 
HERO can be applied to different types of VidL tasks as a multi-channel video encoder. To evaluate on VALUE tasks, we perform task-specific adaptation by adding different task heads.  
See Appendix for more details.

\textbf{Pre-training.} We directly adopt the pre-trained checkpoint released in HERO, which was pre-trained on over 7M video clips from HowTo100M~\cite{miech2019howto100m} and TV dataset~\cite{lei2018tvqa}, with 4 pre-training tasks, \emph{e.g.}, Masked Language Modeling and Video-Subtitle Matching. For finetune-only experiments, model parameters are initialized with pre-trained RoBERTa weights~\cite{liu2019roberta,Wolf2019HuggingFacesTS}.

\vspace{-1mm}
\subsection{Impact of Input Channels and Video-Subtitle Fusion Methods}\label{sec:modality}

In this section, we investigate how information from both video and subtitle channels can be used effectively in multi-channel videos. Specifically, we try to answer the following questions:
 
\textbf{Q1: Is video or subtitle channel alone sufficient to achieve good performance?} Most previous works only leverage visual information from the video channel~\cite{miech2019howto100m,zhu2020actbert}. To assess the importance of the subtitle channel, we evaluate and compare three models: ($i$) video-only, where the model takes only visual features as input;
($ii$) sub-only, where the model takes only subtitle sentences as input;
and ($iii$) video+sub, where the model takes both visual features and subtitle sentences as input.
Results are summarized in Table~\ref{tab:modality}. 
When leveraging both video and subtitle channels, the model achieves the highest meta-average score (52.52) with the best performance across all VALUE tasks. 
 
We also observe that QA tasks generally benefit more from subtitle channel than video channel, but not so for retrieval and captioning tasks. For tasks collected on multi-channel videos (\emph{i.e.}, TV and How2 videos), the model needs to exploit information from both channels to achieve the best performance. For tasks that are originally collected without subtitle channel (\emph{i.e.}, YC2 and VATEX videos), adding subtitle channel still helps. 
Especially for YC2 tasks (YC2R and YC2C), the subtitle-only model performs significantly better than video-only model. 
This is not surprising, as the cooking steps are often clearly described in the dialogues/monologues of cooking videos, so that there is a higher correlation between the retrieval query and the caption. Vice versa, VATEX tasks rely more on video channel than subtitle channel, as the 10-second videos in VATEX focus more on human activities and half of them have no subtitles. 
 
\begin{table}[!t]
\caption{
Impact of \textbf{input channels}. For video-only experiments, we replace all subtitle texts with empty strings. For sub-only experiments, the visual features are replaced with zero vectors. All results are reported on Val/Test (public) split without pre-training.
}
\vspace{2mm}
\tablestyle{2pt}{1.1} 
\def\w{25pt} 
\resizebox{.98\textwidth}{!}{
\begin{tabular}{l|R{\w}R{\w}R{\w}r|R{\w}R{\w}R{\w}R{\w}|R{\w}R{\w}r|R{\w}}
\multirow{2}{*}{ \newcellc{Input\\Channel}} & TVR & How2R & YC2R & \newcellr{VATEX-\\EN-R}&  TVQA   & \newcellr{How2-\\QA} & \newcellr{VIO-\\LIN} & VLEP  & TVC & YC2C & \newcellr{VATEX-\\EN-C}& \multirow{2}{*}{\newcellr{Meta-\\Ave}}\\  
\cline{2-5}\cline{6-9} \cline{10-12}
& AveR & AveR & AveR & AveR & Acc. & Acc.& Acc. & Acc. & C & C & C & \\
\shline
Video-only & 4.49 & 1.70 & 9.74 & 57.50 & 44.17  & 60.42 & 58.53 & 57.56 & 37.52 & 53.61 & 51.14 & 39.67 \\
Sub-only & 1.95 & 0.98 & 32.31 & 5.21 & 70.15  & 68.15 & 66.26 & 58.06 & 38.74 & 93.33 & 9.28 & 40.40\\
Video+Sub & \bf 7.72 & \bf 1.91 & \bf 33.91 & \bf 58.99 & \bf 71.08  & \bf 69.44 & \bf 66.83 & \bf 58.79  & \bf 48.48 & \bf 108.46 & \bf 52.15 & \bf 52.52\\
\end{tabular}
}
\vspace{-2mm}
\label{tab:modality}
\end{table}
 \begin{table}[!t]
\caption{
Impact of \textbf{video-subtitle fusion methods}. Refer to Section~\ref{sec:modality} for detailed explanation of each method. 
HERO's fusion method can also be expressed as \textit{temp. (temporal) align + cross-modal transformer}.
All results are reported on Val/Test (public) split without pre-training.
}
\vspace{2mm}
\tablestyle{2pt}{1.1} 
\def\w{25pt} 
\resizebox{\textwidth}{!}{
\begin{tabular}{p{0.01\textwidth} l|R{\w}R{\w}R{\w}r|R{\w}R{\w}R{\w}R{\w}|R{\w}R{\w}r|R{\w}}
&\multirow{2}{*}{Fusion Method} & TVR & How2R & YC2R & \newcellr{VATEX-\\EN-R}&  TVQA   & \newcellr{How2-\\QA} & \newcellr{VIO-\\LIN} & VLEP  & TVC & YC2C & \newcellr{VATEX-\\EN-C}& \multirow{2}{*}{\newcellr{Meta-\\Ave}}\\  
\cline{3-13}
& & AveR & AveR & AveR & AveR & Acc. & Acc.& Acc. & Acc. & C & C & C & \\
\shline
1 & two-stream & 5.66 & 1.90 & 32.60 & 48.19 & \bf 71.15  & \bf 69.63 & 66.61 & 58.49 & 42.67 & 99.35 & 39.04 & 48.66\\
2 & sequence concat & 5.60 &  2.73 &  \bf 35.55 & \bf 60.24 & 69.61  & 68.99 & 66.09 &  \bf 60.91 & 44.73 & 99.78 & \bf 52.65 & 51.53\\
3 & temp. align + sum & 6.75 & 2.44 & 31.84 & 58.11 & 70.23  & 69.44 & 66.33 &  57.72 & 47.80 & 104.97 & 52.07 & 51.61\\
4 & temp. align + concat & 7.10 & \bf3.19 & 32.59 & 57.33 & 69.81  & 69.31 & 66.16 & 58.54 & 47.12 & 100.90 & 52.09 & 51.29\\
5 & HERO & \bf 7.72 & 1.91 & 33.91 & 58.99 &  71.08  &  69.44 &  \bf 66.83 & 58.79 & \bf 48.48 & \bf 108.46 & 52.15 & \bf 52.52 \\
\end{tabular}
}
\vspace{-2mm}
\label{tab:fusion}
\end{table}
\textbf{Q2: How to effectively fuse video and subtitle embeddings?} 
To answer this, we propose several model variants based on HERO, and compare their performance in Table~\ref{tab:fusion}. 
The simplest baseline is a two-stream architecture~\cite{simonyan2014two,lei2018tvqa}, where the video segments and subtitle sentences are processed separately with different streams to obtain a modality-specific prediction. The final prediction is the average of the predictions from the two streams. 
Such a late fusion method results in the worst performance (meta-average score of 48.66), as the predictions based on the single-channel inputs are independently modeled without considering information from the other channel. 

We further investigate 3 simple ways to fuse video and subtitle embeddings at an earlier stage. 
The three baseline methods are: ($i$) \textit{sequence concat}, concatenating embeddings at sequence level without temporal alignment;
($ii$) \textit{temp. (temporal) align + sum}, summation of the temporally aligned video segment embeddings and subtitle sentence embeddings;
and ($iii$) \textit{temp. align + concat}, concatenation of the temporally aligned video segment embeddings with subtitle sentence embeddings at feature level. 
Finally, we compare with the video-subtitle fusion method proposed in HERO, where the temporally aligned video segments and subtitle sentence tokens are fed into the cross-modal transformer to compute the contextualized embeddings for each video segment. 
These fused embeddings from all the methods above are then fed into the same temporal transformer to learn the global video context and obtain the final video embeddings. 

As shown in Table~\ref{tab:fusion}, HERO achieves the highest meta-average score (52.52), but its performance is sub-optimal on some tasks. For example, the best performance on VATEX tasks is achieved by \textit{sequence concat}, which also outperforms HERO on How2R, YC2R and VLEP. We speculate that the joint video and subtitle representations for those relatively short videos (\emph{e.g.}, VLEP and VATEX) can also be modeled well by simply concatenating VidL embeddings at sequence level, without explicitly aligning them on the temporal domain. Note that the goal is to find a generalizable video-subtitle fusion method that can perform well across 11 VALUE tasks. Therefore, we use HERO as the optimal method for future experiments.

\vspace{-1mm}
\subsection{Impact of Visual Representations}\label{sec:vis_feat}
The common practice to represent a video~\cite{lei2020tvr,sun2019videobert} is to extract 2D appearance features from pre-trained 2D models (\emph{e.g.}, ResNet~\cite{he2016deep}) and 3D motion features from 3D models (\emph{e.g.}, SlowFast~\cite{feichtenhofer2019slowfast}) at the same fixed frame rate, then concatenate them together.
In this section, we investigate the impact of using different visual representations for videos.\footnote{All visual features are released to reproduce the experimental results in this section.}  

We leverage several pre-trained models to extract video features.  For 2D appearance features, we start with the widely adopted ResNet(-152)~\cite{he2016deep} pre-trained on ImageNet~\cite{imagenet}. Recent work~\cite{lei2021less,luo2021clip4clip,bain2021frozen} show that with image-text pre-training, models trained on 2D features alone can achieve decent performance on many video-and-language (VidL) tasks.
Thus, we further evaluate 2D features generated by ViT~\cite{vit2020} in CLIP~\cite{radford2021learning}, which is pre-trained with a large-scale image-text corpus. For 3D motion features, we also evaluate two variants, one from Kinetics~\cite{kay2017kinetics} pre-trained SlowFast~\cite{feichtenhofer2019slowfast} model and the other from an S3D model~\cite{xie2018s3d} pre-trained on 100M video-text pairs~\cite{miech2020end}. 
In addition, we explore different combinations of these features by concatenating 2D features with 3D features following common practice. 
Through this investigation, we hope to understand whether VALUE tasks are designed to favor 2D appearance information from sparsely sampled frames, or require 3D motion information from dense video frames, or rely on both to accomplish these tasks. 
Results are presented in Appendix.
Without pre-training, we found that the best performance is achieved by CLIP-ViT+SlowFast, suggesting that both appearance and motion information are required to handle VALUE tasks.
With pre-training (from HERO's pre-trained checkpoint, trained using ResNet+SlowFast features), ResNet+SlowFast achieves the best performance, likely due to the better matched pre-training and finetuning setting.\footnote{
We did not perform pre-training using other visual features due to its enormous computation cost.
}
In the following, we base all of our experiments on ResNet+SlowFast.

\begin{table*}[t!]\centering
\caption{
\textbf{Task transferability.} We train model on one task and test it on another task of the same task type.
All results are reported on Val/Test (public) split without pre-training. The best and second best performance are highlighted with bold and underline, respectively.
\label{tab:task_transfer_split}
}
\subfloat[Retrieval Tasks.
\label{tab:retrieval_transfer}]{\tablestyle{2pt}{1.0}
\def\w{25pt} 
\scalebox{0.82}{
\begin{tabular}{l|R{20pt}R{25pt}R{25pt}R{28pt}}
Train Data & TVR & How2R & YC2R & \newcellr{VATEX-\\R} \\
\shline
TVR &  \bf 7.72 & \underline{0.00} & 0.35 & 2.79 \\
How2R &  \underline{0.03} &  \bf 1.91 & \underline{10.30} & \underline{10.31}  \\
YC2R& - & - &  \bf 33.91 & 1.01\\
VATEX-R & - & - & 3.82 &  \bf 58.99\\
\end{tabular}
}}
\hspace{1mm}
\subfloat[QA Tasks. \label{tab:qa_transfer}]{\tablestyle{2pt}{1.0}
\def\w{25pt}
\scalebox{0.8}{
\begin{tabular}{l|R{\w}R{\w}R{\w}R{\w}}
Train Data & TVQA   & \newcellr{How2-\\QA}& \newcellr{VIO-\\LIN} & VLEP\\
\shline
TVQA & \bf 71.08  & 36.89 & 50.01 & 53.23 \\
How2QA & 21.75  & \bf 69.44 & \underline{53.85} & \underline{55.65}\\
VIOLIN &20.12  & \underline{40.55}& \bf 66.83 & 44.26 \\
VLEP &  \underline{22.16}  & 26.04 & 50.00 & \bf 58.79\\
\end{tabular}
}}
\hspace{1mm}
\subfloat[Captioning Tasks. 
\label{tab:caption_transfer}]{\tablestyle{2pt}{1.0}
\def\w{25pt}
\scalebox{0.8}{
\begin{tabular}{l|R{20pt}R{25pt}R{28pt}}
Train Data & TVC & YC2C & \newcellr{VATEX-\\C}\\
\shline
TVC & \bf 48.48 & 1.35 & \underline{1.72} \\
YC2C & 0.43 & \bf 108.46 & 0.74\\
VATEX-C & \underline{4.25} & \underline{7.09} & \bf 52.15 \\
\end{tabular}
}
}
\vspace{-3mm}
\end{table*}

\vspace{-1mm}
\subsection{Task Transferability Evaluation}\label{sec:transfer}
In this section, we study how VALUE tasks relate to each other. 
Specifically, we train model on one task and test it on another task of the same type. 
Results are summarized in Table~\ref{tab:task_transfer_split}. Across all task types, the absolutely low performance when transferring the model trained on one task to another indicates that there are significant differences between tasks. The differences can be caused by domain gaps (\emph{e.g.}, TV videos in TVQA and instructional videos in How2QA), discrepancies in video length (\emph{e.g.}, model trained on 60-90 seconds long videos in TVC may not work well on 10-second long videos in VATEX-EN-C) and different task formalization (\emph{e.g.}, model trained on YC2R cannot directly apply to TVR). These results in turn suggest that VALUE supports diverse VidL tasks, thus providing a comprehensive evaluation for VidL understanding systems.

\subsection{Multi-Task Learning Evaluation}\label{sec:multi-task}

\begin{table}[!t]
\caption{
Evaluation of \textbf{multi-task learning baselines} on Test (leaderboard) set.  Results are reported on HERO architecture with ResNet+SlowFast features. We compare the following model training settings: single-task training (ST), multi-task training (MT) by tasks or domains, all-task training (AT) and AT first then ST (AT $\rightarrow$ ST). The best performance (of each block) are highlighted with bold (underline).
}
\vspace{2mm}
\tablestyle{2pt}{1.1} 
\def\w{25pt} 
\resizebox{\textwidth}{!}{
\begin{tabular}{R{0.02\textwidth}l|R{\w}R{\w}R{\w}r|R{\w}R{\w}R{\w}R{\w}|R{\w}R{\w}r|R{\w}}
& \multirow{2}{*}{ \newcelll{Training\\Setting}} & TVR & How2R & YC2R & \newcellr{VATEX-\\EN-R}&  TVQA   & \newcellr{How2-\\QA} & \newcellr{VIO-\\LIN} & VLEP  & TVC & YC2C & \newcellr{VATEX-\\EN-C}& \multirow{2}{*}{\newcellr{Meta-\\Ave}}\\  
\cline{3-13}
& & AveR & AveR & AveR & AveR & Acc. & Acc.& Acc. & Acc. & C & C & C & \\
\shline
1 & Human & - & - & - & - & 89.41 & 90.32  & 91.39  & 90.50 &  62.89 & - & 62.66 & - \\
\shline
& \multicolumn{5}{l}{\textit{Finetune-only}}\\
\hline
2 & ST & 7.70 & 1.74 & 40.69 & 38.34 & 70.54  & 69.00 & 63.75 & 57.94 & \underline{46.76} & \underline{106.24} & \underline{52.16} & 50.44\\
3 & MT by Task  & 7.75 & 1.90 & 46.38 & 38.17 & 71.26  & 71.43 & 64.74 & 68.01 & 46.01 & 105.22 & 51.07 & 52.00 \\
4 & MT by Domain  & 10.01 & \underline{2.69} & 44.58 & 36.10 & 73.94  & 70.01 & \underline{65.93} & 67.37 &46.53 & 100.74 & 50.46  & 51.97\\
5 & AT & 9.76 & 2.42 & 47.91 & 37.33 & \underline{73.98}  & 71.14 & 65.80 & \underline{68.03} & 46.46 & 101.72 & 51.07 & 52.33\\
6 & AT\Arrow{.2cm}ST& \underline{10.43} & 2.68 & \underline{49.48} & \underline{38.58} & 73.46  & \underline{71.88} & 65.73 & 67.80 & 46.12 & 103.73 & 51.87 & \underline{52.89}\\
\shline
& \multicolumn{5}{l}{\textit{Pre-train + Finetune}}\\
\hline
7 & ST & 12.04 & 4.09 & 57.88 & \bf\underline{40.63} & 74.36  & \bf \underline{74.76} & 65.31 & \bf \underline{68.46} & \bf \underline{48.97} & \bf \underline{127.94} & \bf \underline{52.57} & \bf \underline{57.00} \\
8 & MT by Task & \bf \underline{12.63} & \bf \underline{4.66} & \bf \underline{59.20} & 39.97 & 74.56  & 74.40 & 66.34 & 68.11 & 48.02 & 123.40 & 50.49 & 56.53\\
9 & MT by Domain  & 11.53 & 4.03 & 52.14 & 36.97 & 74.54  & 74.08 & 65.92 & 68.06 & 47.23 & 100.29 & 45.95 & 52.79\\
10 & AT & 11.61 & 4.03 & 52.20 & 38.01 & \bf\underline{75.12}  & 73.66 & 66.60 & 68.27 & 46.04 & 109.11 & 49.74 & 54.04\\
11 & AT\Arrow{.2cm}ST& 12.17 & 4.51 & 54.16 & 38.86 & 75.05  & 74.24 & \bf \underline{66.93} & 67.96 & 46.38 & 120.86 & 50.59 & 55.61\\
\end{tabular}
}
\vspace{-2mm}
\label{tab:multi_task}
\end{table}
The low performance observed in the transfer evaluation of task-specifically trained models leads to a natural question: \textit{can one model conquer them all?} In this section, we investigate several multi-task learning baselines and report the results in Table~\ref{tab:multi_task}. We first establish the baseline performance by training single-task models on HERO architecture for each of the 11 datasets (ST, L2). We also include human performance (L1) on eligible QA and captioning tasks.\footnote{See 
Appendix for more information on human evaluation.} Next, we compare different multi-task learning baselines.

\vspace{-1mm}
\textbf{Multi-Task Learning by Task Type.} We begin our investigation with the most intuitive setting - jointly training tasks within the same task type (MT by Task, L3). As the tasks of the same type are typically highly related, this is akin to some data augmentation practice. Note that this corresponds to 3 separate multi-task models - one for each task type. Comparing to ST models (L2), we see that MT by Task achieves +1.56 points improvement on meta-average score (52.00 \textit{vs.} 50.44). 
The increase in meta-average score results from performance improvements on retrieval and QA tasks, with larger improvements on tasks with smaller-scale data (\emph{e.g.}, YC2R and VLEP). 

On captioning tasks, multi-task learning results in a slight performance degradation. 
Note that a single decoder is shared among the three captioning tasks, with task-specific vocabularies combined together. 
This combined vocabulary may introduce more noise than single-task learning when applied to a specific captioning dataset. 
Similar performance decrease is consistently observed in other multi-task learning baselines. For simplicity, we leave out discussions on captioning results.

\textbf{Multi-Task Learning by Domain.} We explore another multi-task learning setting, where we jointly train tasks within the same domain (MT by Domain, L4). We first divide the 11 datasets into 2 domains based on video genre: ($i$) TVR, TVQA, VIOLIN, VLEP and TVC are grouped into TV domain, and ($ii$) the rest of the datasets are grouped into YouTube domain. Note that the videos in TV domain largely overlap among different datasets. However, for datasets in YouTube domain, their videos cover more diverse contents (\emph{e.g.}, YC2 videos focus on cooking while VATEX videos present a wide range of human activities).  Compared with ST (L2), MT by Domain improves by +1.53 on meta-average score (51.97 \textit{vs.} 50.44). In TV domain, model performance improves significantly, suggesting that these different tasks require similar understanding about TV videos. Under YouTube domain, we observe similar improvements on most of the datasets except VATEX-EN-R, where the model seems to be over-fitting to the validation split (Table~\ref{tab:multi_task_val} in Appendix).

\textbf{All-Task Learning.}  We switch to the ``extreme'' multi-task setting - a
single model trained on all 11 datasets (AT, L5). This model outperforms separtely trained ST models (L2) for 8 out of 11 tasks and improve
the meta-average score by +1.89 points (52.33 \textit{vs.} 50.44), while the number of parameters are significantly reduced by approximately 11 times. 
Our AT model also outperforms the other two multi-task baselines (L3-4) on meta-average score despite having fewer parameters. This implies that, despite their diversity, tasks across different task types and domains can benefit from joint training.

\textbf{Multi-Task Learning as Pre-training.} 
Finetuning from a multi-task trained model allows the model to take advantage of the additional, diverse supervision captured during multi-task training. 
Following~\cite{lu201912}, we explore finetuning each task (AT $\rightarrow$ ST, L6) from the multi-task learned weights (L5). 
Results show that this strategy further improves meta-average by +0.56 points (52.89 \textit{vs.} 52.33). 

\textbf{Combining Multi-Task Finetuning with Pre-training.} In L7-11, we take advantage of pre-trained HERO model and repeat experiments in L2-6. When compared with their counterparts without pre-training, we observe consistent performance improvements across all training settings considered. However, pre-training and multi-task finetuning often do not complement each other. Especially on captioning tasks, the performance degradation from multi-task finetuning is even more severe. The best performance is achieved in single-task finetuning with a meta-average score of 57.00 (L7). However, our best model is still far from achieving human parity (L1), especially on QA tasks.

\vspace{-2mm}
\section{Conclusion and Discussion}\label{sec:conclusion}
We introduce VALUE, a comprehensive benchmark for evaluating video-and-language (VidL) understanding systems. VALUE includes 11 VidL datasets with multi-channel video inputs over 3 popular tasks, covering a wide range of video genres, video lengths, task difficulties and data volumes. Through extensive experiments, we conclude that 
designing general-purpose VidL models still remains challenging. We believe that VALUE provides fertile soil for addressing this challenge. For future work, we plan to add diagnostic datasets and support analysis of submitted models both quantitatively and qualitatively, to provide more insights into pushing the state of the art on VALUE.

Although we aim for a comprehensive video-and-language evaluation benchmark, as discussed in Section~\ref{sec:dataset_and_evaluation}, our benchmark currently only contains a selected set of datasets and tasks. It is worthwhile to add more eligible datasets and tasks (considering their diversity, difficulty, etc.) as the next step of the benchmark. Meanwhile, due to the limited availability of multilingual VidL datasets, all datasets covered in the benchmark are of a single language (\textit{i.e.}, English). Future work could consider multilingual VidL datasets~\cite{wang2019vatex,sigurdsson2020visual,huang2021multilingual,lei2021mtvr} as a complementary evaluation to further test systems' ability on processing information in different languages.

\small{
\bibliographystyle{plain}
\bibliography{references} 

\begin{thebibliography}{10}

\bibitem{agirre2015semeval}
Eneko Agirre, Carmen Banea, Claire Cardie, Daniel Cer, Mona Diab, Aitor
  Gonzalez-Agirre, Weiwei Guo, Inigo Lopez-Gazpio, Montse Maritxalar, Rada
  Mihalcea, et~al.
\newblock Semeval-2015 task 2: Semantic textual similarity, english, spanish
  and pilot on interpretability.
\newblock In {\em Proceedings of the 9th international workshop on semantic
  evaluation (SemEval 2015)}, 2015.

\bibitem{agirre2014semeval}
Eneko Agirre, Carmen Banea, Claire Cardie, Daniel Cer, Mona Diab, Aitor
  Gonzalez-Agirre, Weiwei Guo, Rada Mihalcea, German Rigau, and Janyce Wiebe.
\newblock Semeval-2014 task 10: Multilingual semantic textual similarity.
\newblock In {\em Proceedings of the 8th international workshop on semantic
  evaluation (SemEval 2014)}, 2014.

\bibitem{agirre2016semeval}
Eneko Agirre, Carmen Banea, Daniel Cer, Mona Diab, Aitor Gonzalez~Agirre, Rada
  Mihalcea, German Rigau~Claramunt, and Janyce Wiebe.
\newblock Semeval-2016 task 1: Semantic textual similarity, monolingual and
  cross-lingual evaluation.
\newblock In {\em SemEval-2016. 10th International Workshop on Semantic
  Evaluation}, 2016.

\bibitem{agirre2012semeval}
Eneko Agirre, Daniel Cer, Mona Diab, and Aitor Gonzalez-Agirre.
\newblock Semeval-2012 task 6: A pilot on semantic textual similarity.
\newblock In {\em * SEM 2012: The First Joint Conference on Lexical and
  Computational Semantics--Volume 1: Proceedings of the main conference and the
  shared task, and Volume 2: Proceedings of the Sixth International Workshop on
  Semantic Evaluation (SemEval 2012)}, 2012.

\bibitem{agirre2013sem}
Eneko Agirre, Daniel Cer, Mona Diab, Aitor Gonzalez-Agirre, and Weiwei Guo.
\newblock * sem 2013 shared task: Semantic textual similarity.
\newblock In {\em Second joint conference on lexical and computational
  semantics (* SEM), volume 1: proceedings of the Main conference and the
  shared task: semantic textual similarity}, 2013.

\bibitem{albanie2020end}
Samuel Albanie, Yang Liu, Arsha Nagrani, Antoine Miech, Ernesto Coto, Ivan
  Laptev, Rahul Sukthankar, Bernard Ghanem, Andrew Zisserman, Valentin Gabeur,
  et~al.
\newblock The end-of-end-to-end: A video understanding pentathlon challenge
  (2020).
\newblock {\em arXiv preprint arXiv:2008.00744}, 2020.

\bibitem{anne2017localizing}
Lisa Anne~Hendricks, Oliver Wang, Eli Shechtman, Josef Sivic, Trevor Darrell,
  and Bryan Russell.
\newblock Localizing moments in video with natural language.
\newblock In {\em CVPR}, 2017.

\bibitem{didemo}
Lisa Anne~Hendricks, Oliver Wang, Eli Shechtman, Josef Sivic, Trevor Darrell,
  and Bryan Russell.
\newblock Localizing moments in video with natural language.
\newblock In {\em ICCV}, 2017.

\bibitem{bain2021frozen}
Max Bain, Arsha Nagrani, G{\"u}l Varol, and Andrew Zisserman.
\newblock Frozen in time: A joint video and image encoder for end-to-end
  retrieval.
\newblock {\em arXiv preprint arXiv:2104.00650}, 2021.

\bibitem{bowman2015large}
Samuel~R Bowman, Gabor Angeli, Christopher Potts, and Christopher~D Manning.
\newblock A large annotated corpus for learning natural language inference.
\newblock In {\em EMNLP}, 2015.

\bibitem{msvd}
David Chen and William Dolan.
\newblock Collecting highly parallel data for paraphrase evaluation.
\newblock In {\em ACL}, 2011.

\bibitem{conneau2018senteval}
Alexis Conneau and Douwe Kiela.
\newblock Senteval: An evaluation toolkit for universal sentence
  representations.
\newblock In {\em LREC}, 2018.

\bibitem{imagenet}
J.~Deng, W.~Dong, R.~Socher, L.-J. Li, K.~Li, and L.~Fei-Fei.
\newblock {ImageNet: A Large-Scale Hierarchical Image Database}.
\newblock In {\em CVPR}, 2009.

\bibitem{deng2009imagenet}
Jia Deng, Wei Dong, Richard Socher, Li-Jia Li, Kai Li, and Li~Fei-Fei.
\newblock Imagenet: A large-scale hierarchical image database.
\newblock In {\em CVPR}, 2009.

\bibitem{denkowski2014meteor}
Michael Denkowski and Alon Lavie.
\newblock Meteor universal: Language specific translation evaluation for any
  target language.
\newblock In {\em Proceedings of the ninth workshop on statistical machine
  translation}, 2014.

\bibitem{devlin2018bert}
Jacob Devlin, Ming-Wei Chang, Kenton Lee, and Kristina Toutanova.
\newblock Bert: Pre-training of deep bidirectional transformers for language
  understanding.
\newblock In {\em NAACL}, 2019.

\bibitem{vit2020}
Alexey Dosovitskiy, Lucas Beyer, Alexander Kolesnikov, Dirk Weissenborn,
  Xiaohua Zhai, Thomas Unterthiner, Mostafa Dehghani, Matthias Minderer, Georg
  Heigold, Sylvain Gelly, et~al.
\newblock An image is worth 16x16 words: Transformers for image recognition at
  scale.
\newblock In {\em ICLR}, 2020.

\bibitem{fang2020filter}
Yuwei Fang, Shuohang Wang, Zhe Gan, Siqi Sun, and Jingjing Liu.
\newblock Filter: An enhanced fusion method for cross-lingual language
  understanding.
\newblock In {\em AAAI}, 2021.

\bibitem{feichtenhofer2019slowfast}
Christoph Feichtenhofer, Haoqi Fan, Jitendra Malik, and Kaiming He.
\newblock Slowfast networks for video recognition.
\newblock In {\em ICCV}, 2019.

\bibitem{gabeur2020multi}
Valentin Gabeur, Chen Sun, Karteek Alahari, and Cordelia Schmid.
\newblock Multi-modal transformer for video retrieval.
\newblock In {\em ECCV}, 2020.

\bibitem{gao2017tall}
Jiyang Gao, Chen Sun, Zhenheng Yang, and Ram Nevatia.
\newblock Tall: Temporal activity localization via language query.
\newblock In {\em CVPR}, 2017.

\bibitem{he2016deep}
Kaiming He, Xiangyu Zhang, Shaoqing Ren, and Jian Sun.
\newblock Deep residual learning for image recognition.
\newblock In {\em CVPR}, 2016.

\bibitem{hu2020xtreme}
Junjie Hu, Sebastian Ruder, Aditya Siddhant, Graham Neubig, Orhan Firat, and
  Melvin Johnson.
\newblock Xtreme: A massively multilingual multi-task benchmark for evaluating
  cross-lingual generalisation.
\newblock In {\em ICML}, 2020.

\bibitem{huang2021multilingual}
Po-Yao Huang, Mandela Patrick, Junjie Hu, Graham Neubig, Florian Metze, and
  Alexander Hauptmann.
\newblock Multilingual multimodal pre-training for zero-shot cross-lingual
  transfer of vision-language models.
\newblock {\em arXiv preprint arXiv:2103.08849}, 2021.

\bibitem{jang2017tgif}
Yunseok Jang, Yale Song, Youngjae Yu, Youngjin Kim, and Gunhee Kim.
\newblock Tgif-qa: Toward spatio-temporal reasoning in visual question
  answering.
\newblock In {\em CVPR}, 2017.

\bibitem{jiang2020divide}
Jianwen Jiang, Ziqiang Chen, Haojie Lin, Xibin Zhao, and Yue Gao.
\newblock Divide and conquer: Question-guided spatio-temporal contextual
  attention for video question answering.
\newblock In {\em AAAI}, 2020.

\bibitem{kay2017kinetics}
Will Kay, Joao Carreira, Karen Simonyan, Brian Zhang, Chloe Hillier, Sudheendra
  Vijayanarasimhan, Fabio Viola, Tim Green, Trevor Back, Paul Natsev, et~al.
\newblock The kinetics human action video dataset.
\newblock {\em arXiv preprint arXiv:1705.06950}, 2017.

\bibitem{krishna2017dense}
Ranjay Krishna, Kenji Hata, Frederic Ren, Li~Fei-Fei, and Juan Carlos~Niebles.
\newblock Dense-captioning events in videos.
\newblock In {\em ICCV}, 2017.

\bibitem{le2020hierarchical}
Thao~Minh Le, Vuong Le, Svetha Venkatesh, and Truyen Tran.
\newblock Hierarchical conditional relation networks for video question
  answering.
\newblock In {\em CVPR}, 2020.

\bibitem{lei2021mtvr}
Jie Lei, Tamara~L Berg, and Mohit Bansal.
\newblock mtvr: Multilingual moment retrieval in videos.
\newblock In {\em ACL}, 2021.

\bibitem{lei2021less}
Jie Lei, Linjie Li, Luowei Zhou, Zhe Gan, Tamara~L. Berg, Mohit Bansal, and
  Jingjing Liu.
\newblock Less is more: Clipbert for video-and-language learningvia sparse
  sampling.
\newblock In {\em CVPR}, 2021.

\bibitem{lei2018tvqa}
Jie Lei, Licheng Yu, Mohit Bansal, and Tamara Berg.
\newblock Tvqa: Localized, compositional video question answering.
\newblock In {\em EMNLP}, 2018.

\bibitem{lei2019tvqa+}
Jie Lei, Licheng Yu, Tamara Berg, and Mohit Bansal.
\newblock Tvqa+: Spatio-temporal grounding for video question answering.
\newblock In {\em ACL}, 2020.

\bibitem{lei2020more}
Jie Lei, Licheng Yu, Tamara Berg, and Mohit Bansal.
\newblock What is more likely to happen next? video-and-language future event
  prediction.
\newblock In {\em EMNLP}, 2020.

\bibitem{lei2020tvr}
Jie Lei, Licheng Yu, Tamara~L Berg, and Mohit Bansal.
\newblock Tvr: A large-scale dataset for video-subtitle moment retrieval.
\newblock In {\em ECCV}, 2020.

\bibitem{li2020hero}
Linjie Li, Yen-Chun Chen, Yu~Cheng, Zhe Gan, Licheng Yu, and Jingjing Liu.
\newblock Hero: Hierarchical encoder for video+ language omni-representation
  pre-training.
\newblock In {\em EMNLP}, 2020.

\bibitem{liang2020xglue}
Yaobo Liang, Nan Duan, Yeyun Gong, Ning Wu, Fenfei Guo, Weizhen Qi, Ming Gong,
  Linjun Shou, Daxin Jiang, Guihong Cao, et~al.
\newblock Xglue: A new benchmark dataset for cross-lingual pre-training,
  understanding and generation.
\newblock In {\em EMNLP}, 2020.

\bibitem{linrouge}
Chin-Yew Lin.
\newblock Rouge: A package for automatic evaluation of summaries.
\newblock In {\em ACL}, 2004.

\bibitem{lin2021vx2text}
Xudong Lin, Gedas Bertasius, Jue Wang, Shih-Fu Chang, Devi Parikh, and Lorenzo
  Torresani.
\newblock Vx2text: End-to-end learning of video-based text generation from
  multimodal inputs.
\newblock In {\em CVPR}, 2021.

\bibitem{liu2020violin}
Jingzhou Liu, Wenhu Chen, Yu~Cheng, Zhe Gan, Licheng Yu, Yiming Yang, and
  Jingjing Liu.
\newblock Violin: A large-scale dataset for video-and-language inference.
\newblock In {\em CVPR}, 2020.

\bibitem{liu2019use}
Yang Liu, Samuel Albanie, Arsha Nagrani, and Andrew Zisserman.
\newblock Use what you have: Video retrieval using representations from
  collaborative experts.
\newblock In {\em BMVC}, 2020.

\bibitem{liu2019roberta}
Yinhan Liu, Myle Ott, Naman Goyal, Jingfei Du, Mandar Joshi, Danqi Chen, Omer
  Levy, Mike Lewis, Luke Zettlemoyer, and Veselin Stoyanov.
\newblock Roberta: A robustly optimized bert pretraining approach.
\newblock {\em arXiv preprint arXiv:1907.11692}, 2019.

\bibitem{AdamW}
Ilya Loshchilov and Frank Hutter.
\newblock Decoupled weight decay regularization.
\newblock In {\em ICLR}, 2019.

\bibitem{lu201912}
Jiasen Lu, Vedanuj Goswami, Marcus Rohrbach, Devi Parikh, and Stefan Lee.
\newblock 12-in-1: Multi-task vision and language representation learning.
\newblock In {\em CVPR}, 2020.

\bibitem{luo2021clip4clip}
Huaishao Luo, Lei Ji, Ming Zhong, Yang Chen, Wen Lei, Nan Duan, and Tianrui Li.
\newblock Clip4clip: An empirical study of clip for end to end video clip
  retrieval.
\newblock {\em arXiv preprint arXiv:2104.08860}, 2021.

\bibitem{miech2020end}
Antoine Miech, Jean-Baptiste Alayrac, Lucas Smaira, Ivan Laptev, Josef Sivic,
  and Andrew Zisserman.
\newblock End-to-end learning of visual representations from uncurated
  instructional videos.
\newblock In {\em CVPR}, 2020.

\bibitem{miech2019howto100m}
Antoine Miech, Dimitri Zhukov, Jean-Baptiste Alayrac, Makarand Tapaswi, Ivan
  Laptev, and Josef Sivic.
\newblock Howto100m: Learning a text-video embedding by watching hundred
  million narrated video clips.
\newblock In {\em ICCV}, 2019.

\bibitem{ott2018scaling}
Myle Ott, Sergey Edunov, David Grangier, and Michael Auli.
\newblock Scaling neural machine translation.
\newblock {\em arXiv preprint arXiv:1806.00187}, 2018.

\bibitem{ouyang2020ernie}
Xuan Ouyang, Shuohuan Wang, Chao Pang, Yu~Sun, Hao Tian, Hua Wu, and Haifeng
  Wang.
\newblock Ernie-m: Enhanced multilingual representation by aligning
  cross-lingual semantics with monolingual corpora.
\newblock {\em arXiv preprint arXiv:2012.15674}, 2020.

\bibitem{pang2005seeing}
Bo~Pang and Lillian Lee.
\newblock Seeing stars: Exploiting class relationships for sentiment
  categorization with respect to rating scales.
\newblock In {\em ACL}, 2005.

\bibitem{papineni2002bleu}
Kishore Papineni, Salim Roukos, Todd Ward, and Wei-Jing Zhu.
\newblock Bleu: a method for automatic evaluation of machine translation.
\newblock In {\em ACL}, 2002.

\bibitem{paszke2017automatic}
Adam Paszke, Sam Gross, Soumith Chintala, Gregory Chanan, Edward Yang, Zachary
  DeVito, Zeming Lin, Alban Desmaison, Luca Antiga, and Adam Lerer.
\newblock Automatic differentiation in pytorch.
\newblock 2017.

\bibitem{patrick2020support}
Mandela Patrick, Po-Yao Huang, Yuki Asano, Florian Metze, Alexander Hauptmann,
  Jo{\~a}o Henriques, and Andrea Vedaldi.
\newblock Support-set bottlenecks for video-text representation learning.
\newblock In {\em ICLR}, 2021.

\bibitem{radford2021learning}
Alec Radford, Jong~Wook Kim, Chris Hallacy, Aditya Ramesh, Gabriel Goh,
  Sandhini Agarwal, Girish Sastry, Amanda Askell, Pamela Mishkin, Jack Clark,
  et~al.
\newblock Learning transferable visual models from natural language
  supervision.
\newblock {\em arXiv preprint arXiv:2103.00020}, 2021.

\bibitem{radford2018improving}
Alec Radford, Karthik Narasimhan, Tim Salimans, and Ilya Sutskever.
\newblock Improving language understanding by generative pre-training.
\newblock 2018.

\bibitem{rohrbach2015dataset}
Anna Rohrbach, Marcus Rohrbach, Niket Tandon, and Bernt Schiele.
\newblock A dataset for movie description.
\newblock In {\em CVPR}, 2015.

\bibitem{sigurdsson2020visual}
Gunnar~A Sigurdsson, Jean-Baptiste Alayrac, Aida Nematzadeh, Lucas Smaira,
  Mateusz Malinowski, Jo{\~a}o Carreira, Phil Blunsom, and Andrew Zisserman.
\newblock Visual grounding in video for unsupervised word translation.
\newblock In {\em CVPR}, 2020.

\bibitem{simonyan2014two}
Karen Simonyan and Andrew Zisserman.
\newblock Two-stream convolutional networks for action recognition in videos.
\newblock In {\em NeurIPS}, 2014.

\bibitem{socher2013recursive}
Richard Socher, Alex Perelygin, Jean Wu, Jason Chuang, Christopher~D Manning,
  Andrew~Y Ng, and Christopher Potts.
\newblock Recursive deep models for semantic compositionality over a sentiment
  treebank.
\newblock In {\em EMNLP}, 2013.

\bibitem{sun2019videobert}
Chen Sun, Austin Myers, Carl Vondrick, Kevin Murphy, and Cordelia Schmid.
\newblock Videobert: A joint model for video and language representation
  learning.
\newblock In {\em ICCV}, 2019.

\bibitem{tang2021decembert}
Zineng Tang, Jie Lei, and Mohit Bansal.
\newblock Decembert: Learning from noisy instructional videos via dense
  captions and entropy minimization.
\newblock In {\em NAACL}, 2021.

\bibitem{vaswani2017attention}
Ashish Vaswani, Noam Shazeer, Niki Parmar, Jakob Uszkoreit, Llion Jones,
  Aidan~N Gomez, {\L}ukasz Kaiser, and Illia Polosukhin.
\newblock Attention is all you need.
\newblock In {\em NeurIPS}, 2017.

\bibitem{vedantam2015cider}
Ramakrishna Vedantam, C~Lawrence~Zitnick, and Devi Parikh.
\newblock Cider: Consensus-based image description evaluation.
\newblock In {\em CVPR}, 2015.

\bibitem{wang2019superglue}
Alex Wang, Yada Pruksachatkun, Nikita Nangia, Amanpreet Singh, Julian Michael,
  Felix Hill, Omer Levy, and Samuel~R Bowman.
\newblock Superglue: A stickier benchmark for general-purpose language
  understanding systems.
\newblock In {\em NeurIPS}, 2019.

\bibitem{wang2018glue}
Alex Wang, Amanpreet Singh, Julian Michael, Felix Hill, Omer Levy, and Samuel~R
  Bowman.
\newblock Glue: A multi-task benchmark and analysis platform for natural
  language understanding.
\newblock In {\em ICLR}, 2019.

\bibitem{wang2019vatex}
Xin Wang, Jiawei Wu, Junkun Chen, Lei Li, Yuan-Fang Wang, and William~Yang
  Wang.
\newblock Vatex: A large-scale, high-quality multilingual dataset for
  video-and-language research.
\newblock In {\em ICCV}, 2019.

\bibitem{wei2020learning}
Xiangpeng Wei, Rongxiang Weng, Yue Hu, Luxi Xing, Heng Yu, and Weihua Luo.
\newblock On learning universal representations across languages.
\newblock In {\em ICLR}, 2021.

\bibitem{Wolf2019HuggingFacesTS}
Thomas Wolf, Julien Chaumond, Lysandre Debut, Victor Sanh, Clement Delangue,
  Anthony Moi, Pierric Cistac, Morgan Funtowicz, Joe Davison, Sam Shleifer,
  et~al.
\newblock Transformers: State-of-the-art natural language processing.
\newblock In {\em EMNLP: System Demonstrations}, 2020.

\bibitem{wu2016google}
Yonghui Wu, Mike Schuster, Zhifeng Chen, Quoc~V Le, Mohammad Norouzi, Wolfgang
  Macherey, Maxim Krikun, Yuan Cao, Qin Gao, Klaus Macherey, et~al.
\newblock Google's neural machine translation system: Bridging the gap between
  human and machine translation.
\newblock {\em arXiv preprint arXiv:1609.08144}, 2016.

\bibitem{xie2018s3d}
Saining Xie, Chen Sun, Jonathan Huang, Zhuowen Tu, and Kevin Murphy.
\newblock Rethinking spatiotemporal feature learning: Speed-accuracy trade-offs
  in video classification.
\newblock In {\em ECCV}, 2018.

\bibitem{xu2017video}
Dejing Xu, Zhou Zhao, Jun Xiao, Fei Wu, Hanwang Zhang, Xiangnan He, and Yueting
  Zhuang.
\newblock Video question answering via gradually refined attention over
  appearance and motion.
\newblock In {\em ACM MM}, 2017.

\bibitem{msrvtt-qa}
Dejing Xu, Zhou Zhao, Jun Xiao, Fei Wu, Hanwang Zhang, Xiangnan He, and Yueting
  Zhuang.
\newblock Video question answering via gradually refined attention over
  appearance and motion.
\newblock In {\em ACM MM}, 2017.

\bibitem{xu2016msr}
Jun Xu, Tao Mei, Ting Yao, and Yong Rui.
\newblock Msr-vtt: A large video description dataset for bridging video and
  language.
\newblock In {\em CVPR}, 2016.

\bibitem{xue2020mt5}
Linting Xue, Noah Constant, Adam Roberts, Mihir Kale, Rami Al-Rfou, Aditya
  Siddhant, Aditya Barua, and Colin Raffel.
\newblock mt5: A massively multilingual pre-trained text-to-text transformer.
\newblock In {\em NAACL}, 2020.

\bibitem{yang2019xlnet}
Zhilin Yang, Zihang Dai, Yiming Yang, Jaime Carbonell, Russ~R Salakhutdinov,
  and Quoc~V Le.
\newblock Xlnet: Generalized autoregressive pretraining for language
  understanding.
\newblock In {\em NeurIPS}, 2019.

\bibitem{yu2018mattnet}
Licheng Yu, Zhe Lin, Xiaohui Shen, Jimei Yang, Xin Lu, Mohit Bansal, and
  Tamara~L Berg.
\newblock Mattnet: Modular attention network for referring expression
  comprehension.
\newblock In {\em CVPR}, 2018.

\bibitem{zhou2021cupid}
Luowei Zhou, Jingjing Liu, Yu~Cheng, Zhe Gan, and Lei Zhang.
\newblock Cupid: Adaptive curation of pre-training data for video-and-language
  representation learning.
\newblock {\em arXiv preprint arXiv:2104.00285}, 2021.

\bibitem{zhou2018towards}
Luowei Zhou, Chenliang Xu, and Jason~J Corso.
\newblock Towards automatic learning of procedures from web instructional
  videos.
\newblock In {\em AAAI}, 2018.

\bibitem{zhou2018end}
Luowei Zhou, Yingbo Zhou, Jason~J Corso, Richard Socher, and Caiming Xiong.
\newblock End-to-end dense video captioning with masked transformer.
\newblock In {\em CVPR}, 2018.

\bibitem{zhu2020actbert}
Linchao Zhu and Yi~Yang.
\newblock Actbert: Learning global-local video-text representations.
\newblock In {\em CVPR}, 2020.

\end{thebibliography}
}

\clearpage
\appendix
\section{Additional Data Statistics}\label{sec:vis_data_statistics}
\begin{figure}[h]
    \centering
    \includegraphics[width=\textwidth]{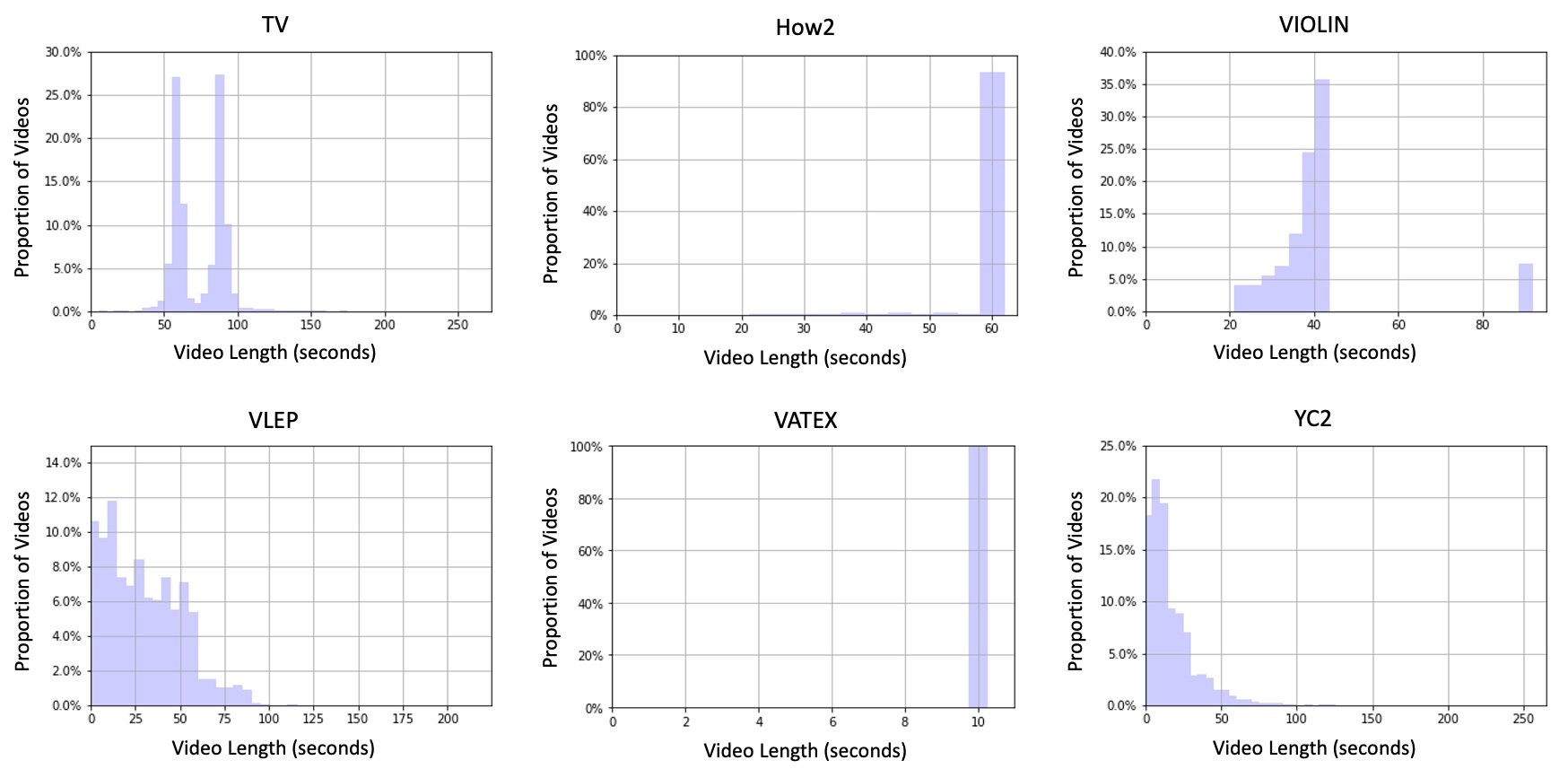}
    \caption{Visualization of video length distribution.}
    \label{fig:video_length}
\end{figure}

We visualize video length distribution for each video data in Figure~\ref{fig:video_length}. Table~\ref{tab:keyword_sub} and~\ref{tab:keyword_ann} summarize the top-20 most frequent nouns and verbs in subtitles and annotations.

\begin{table*}[h]
\caption{
Top-20 most frequent nouns and verbs in subtitles. Character names have been filtered out from TV/VIOLIN/VLEP.
}

\tablestyle{2pt}{1.8} 
\def\w{175pt} 
\resizebox{\textwidth}{!}{
\begin{tabular}{l|L{\w}L{\w}}
Dataset & Nouns & Verbs\\  
\shline
TV & time, something, guy, way, right, man, thing, look, night, anything, god, someone, nothing, life, day, thank, kind, wait, woman, everything &
know, go, think, want, see, need, tell, say, take, make, let, mean, come, find, give, look, talk, believe, love, feel \\
\hline
How2 & bit, way, kind, time, water, today, thing, lot, side, video, music, right, let, something, oil, cup, sugar, top, part, half &
make, want, go, see, know, use, take, put, need, add, let, show, think, start, give, look, keep, cut, come, say \\
\hline
VIOLIN & time, something, god, look, thing, way, right, man, guy, day, night, mom, anything, thank, kind, nothing, wait, life, baby, let & 
know, go, think, want, see, say, tell, take, need, let, come, make, mean, look, give, love, feel, talk, believe, call \\
\hline
VLEP & time, look, kind, something, thing, right, way, man, day, music, lot, everything, let, bit, guy, thank, food, night, place, god &
know, go, think, see, want, need, make, take, let, say, come, love, look, mean, tell, give, try, feel, find, eat\\
\hline
VATEX & way, time, bit, side, look, god, job, alright, thing, kind, hand, man, today, video, day, right, lot, water, thank, something &
go, know, see, want, make, take, think, let, put, say, come, need, keep, look, use, give, start, show, hold, love\\
\hline
YC2 & bit, oil, water, salt, sauce, pepper, time, kind, teaspoon, heat, pan, cup, chicken, onion, half, way, butter, garlic, side, flavor &
add, want, make, put, use, go, take, let, see, cook, know, need, give, mix, start, cut, keep, turn, think, look\\
\end{tabular}
}
\vspace{-2mm}
\label{tab:keyword_sub}
\end{table*}

\begin{table*}[h]
\caption{
Top-20 most frequent nouns and verbs in annotations (query/question/caption). Character names have been filtered out from TVR/TVC/TVQA/VIOLIN/VLEP.
}

\tablestyle{2pt}{1.8} 
\def\w{165pt} 
\resizebox{\textwidth}{!}{
\begin{tabular}{l|L{\w}L{\w}}
Dataset & Nouns & Verbs\\  
\shline
TVR/TVC & walk, hand, room, door, table, conversation, patient, man, phone, woman, apartment, head, talk, bed, front, chair, coffee, arm, tell, couch 
& talking, walk, take, look, put, sitting, stand, turn, open, holding, sits, tell, hold, pick, asks, say, standing, give, go, looking\\
\hline
TVQA & room, patient, hand, door, color, apartment, table, phone, man, office, doctor, woman, shirt, something, couch, hospital, coffee, time, friend, guy 
& say, talking, said, tell, sitting, holding, asked, go, told, asks, going, walk, walked, want, wearing, talk, looking, come, give, standing\\
\hline
How2R & man, woman, video, lady, person, car, bowl, paper, hand, ingredient, girl, food, piece, plant, water, pan, glass, guy, kitchen, chef
& make, explain, talking, show, using, showing, making, explaining, put, add, explains, cut, shown, cooking, describes, cutting, hold, holding, use, take\\
\hline
How2QA & video, color, man, woman, person, lady, hand, name, kind, type, bowl, food, ingredient, shirt, car, girl, item, table, boy, paper
& used, shown, talking, using, put, wearing, added, make, holding, use, seen, cut, hold, want, mentioned, making, add, show, happen, explain\\
\hline
VIOLIN & man, woman, shirt, suit, hair, jacket, blonde, girl, lady, brunette, dress, boy, sweater, grey, friend, room, blue, men, pink, guy
& wearing, tell, want, asks, sitting, say, explains, talking, trying, haired, see, go, think, make, walk, take, know, holding, going, look\\
\hline
VLEP & man, food, door, tell, room, woman, hand, patient, table, phone, shirt, walk, apartment, something, friend, vlogger, baby, girl, question, someone
& say, tell, take, go, put, asks, look, give, start, walk, make, talk, going, leave, want, open, see, eat, turn, continue\\
\hline
VATEX-EN-R/-C & man, woman, person, people, boy, girl, group, hand, someone, music, child, ball, men, baby, water, room, piece, front, kid, floor & playing, using, sitting, play, holding, talking, standing, showing, make, wearing, shown, dancing, riding, put, show, stand, demonstrating, throw, demonstrates, hold\\
\hline
YC2R/YC2C & pan, oil, onion, salt, water, sauce, pepper, bowl, place, mix, pot, egg, potato, stir, chicken, mixture, slice, powder, butter, heat 
& add, cut, mix, put, cook, chopped, remove, fry, take, chop, cover, spread, serve, stir, pour, roll, drain, flip, blend, baking\\
\end{tabular}
}

\vspace{-2mm}
\label{tab:keyword_ann}
\end{table*}

\section{Additional Results}
\subsection{Impact of Visual Representations}\label{app:vis_feat}
Table~\ref{tab:vis_feat} shows the results using different visual representations. Our key observations are summarized as follows: ($i$) We confirm that image-text pre-trained CLIP-ViT features are generalizable to video-and-language (VidL) tasks (L2 vs. L1/3/4). CLIP-ViT features lead to stronger performance than other 2D or 3D features. ($ii$) VALUE tasks also benefit from video-text pre-trained S3D features (L3 vs. L4). However, the performance improvement mostly comes from YC2 tasks, the videos of which are similar to the videos used to pre-train the S3D model. These results imply that the video domain of pre-training data is critical to downstream performance. ($iii$) When taking advantage of both 2D and 3D features, the model achieves the best performance (CLIP-ViT+SlowFast, L7), suggesting that both appearance and motion information are required to solve VALUE tasks. ($iv$) However, 2D and 3D features do not always complement each other. For example, performance on ResNet+S3D (L6) is worse than that on S3D alone (L4). ($v$) Retrieval and captioning tasks greatly depend on the quality of visual representations, while QA performance stays relatively stable with different features. This result agrees with our observations in Table 3 in the main text, where we show that QA tasks rely more on information from subtitle channel.

In addition, we finetune the pre-trained HERO on different visual representations (L9-12). Comparing their counterparts without pre-training, we observe a consistent performance improvement in almost all tasks across all visual representations examined. 
Note that in L10-12, though the video features used in pre-training is different from that used in finetuning, we still observe significant performance gains compared to models without pre-training.
This suggests that the video-language alignment learned via HERO pre-training is transferrable to different visual representations.

\begin{table}[!t]
\caption{
Impact of \textbf{visual representations}. ResNet(-152)~\cite{he2016deep} and SlowFast~\cite{feichtenhofer2019slowfast} are pre-trained on ImageNet~\cite{imagenet} and Kinetics~\cite{kay2017kinetics}, respectively. S3D~\cite{xie2018s3d,miech2020end} is pre-trained with video-text pairs in HowTo100M~\cite{miech2019howto100m}, OpenAI CLIP ViT~\cite{vit2020,radford2021learning} is pre-trained with image-text pairs~\cite{radford2021learning}.  All results are reported on Val/Test (public) split. The best performance (of each block) are highlighted with bold (underline).
}
\vspace{2mm}
\tablestyle{2pt}{1.1} 
\def\w{25pt} 
\resizebox{\textwidth}{!}{
\begin{tabular}{R{0.02\textwidth} l|R{\w}R{\w}R{\w}r|R{\w}R{\w}R{\w}R{\w}|R{\w}R{\w}r|R{\w}}
& \multirow{2}{*}{ Visual Feature} & TVR & How2R & YC2R & \newcellr{VATEX-\\EN-R}&  TVQA   & \newcellr{How2-\\QA} & \newcellr{VIO-\\LIN} & VLEP  & TVC & YC2C & \newcellr{VATEX-\\EN-C}& \multirow{2}{*}{\newcellr{Meta-\\Ave}}\\  
\cline{3-13}
& & AveR & AveR & AveR & AveR & Acc. & Acc.& Acc. & Acc. & C & C & C & \\
\shline
& \multicolumn{5}{l}{\textit{2D Features, Finetune-only}}\\
\hline
1 & ResNet~\cite{he2016deep} & 4.82 & 0.75 & 33.96 & 43.93 & 70.73  & 68.41 & 66.28 & 57.47 & 45.54 & 100.89 & 38.41 & 48.29 \\
2 & CLIP-ViT~\cite{vit2020,radford2021learning} & \underline{7.93} & \underline{1.52} & \underline{35.93} & \underline{62.87} & \underline{71.07}  & \underline{69.34} & \underline{66.80} & \underline{58.27} & \underline{48.99} & \underline{112.25} & \underline{52.42} & \underline{53.40} \\
\hline
& \multicolumn{5}{l}{\textit{3D Features, Finetune-only}}\\
\hline
3 & SlowFast~\cite{feichtenhofer2019slowfast} & 4.71 & \underline{3.19} & 34.82 & \underline{56.19} & 71.13  & 68.31 & 66.00 & 58.11 & \underline{47.77} & 105.85 & \underline{51.20} & 51.57 \\
4 & S3D~\cite{xie2018s3d,miech2020end} & \underline{6.14} & 2.52 & \underline{41.66} & 49.28 & \underline{71.34}  & \underline{69.47} & \underline{66.41} & \underline{58.22} & 47.32 & \bf\underline{125.58} & 42.65 & \underline{52.78} \\
\hline
& \multicolumn{5}{l}{\textit{2D+3D Features, Finetune-only}}\\
\hline
5 & ResNet+SlowFast & 7.72 & 1.91 & 33.91 & 58.99 & 71.08 & 69.44 & 66.83  & 58.79 & 48.48 & 108.46 & 52.15 & 52.52\\
6 & ResNet+S3D & 5.16 & 2.32 & 33.88 & 46.19 & 70.70 & 66.68 & 68.60 & 58.65 & 45.22 & 105.83 & 39.51 & 49.34\\
7 & CLIP-ViT+SlowFast & \underline{8.84} & \underline{2.39} & 34.63 & \underline{65.62} & \underline{71.64}  & \underline{70.21} & \underline{67.21} & 57.56 & \underline{51.47} & \underline{113.23} & \bf \underline{56.97} & \underline{54.52}\\
8 & CLIP-ViT+S3D & 6.66 & 2.27 & \underline{36.68} & 62.35 & 70.27  & 68.54 & 67.06 & \underline{59.13} & 50.05 & 110.18 & 52.77 & 53.27 \\
\shline
& \multicolumn{8}{l}{\textit{2D+3D Features, Pre-train on ResNet+SlowFast then Finetune}}\\
\hline
9 & ResNet+SlowFast & 11.66 & \bf \underline{5.97}  & 48.86 & 61.66 & \bf \underline{74.80}  & \bf \underline{74.32} & \bf \underline{68.98} & \bf \underline{67.40} & 50.46 & \underline{121.89} & 52.58 & \bf \underline{58.05}\\
10 & ResNet+S3D & 9.45 & 5.20 & 47.81 & 47.00  & 72.65  & 72.68 & 67.71 & 65.94 & 46.42 & 117.11 & 38.77 & 53.70\\
11 & CLIP-ViT+SlowFast & \bf \underline{12.92} & 5.02 & 47.81 & \bf \underline{66.49} & 74.25  & 72.87 & 68.33 & 65.60 & \bf \underline{51.56} & 115.83 &
\underline{56.19} & 57.90\\
12 & CLIP-ViT+S3D & 11.85  & 5.43 & \bf \underline{49.52} & 63.37 & 72.56  & 73.64 & 67.92 & 65.82 & 50.26 & 117.58 & 50.73 & 57.15\\
\end{tabular}
}

\label{tab:vis_feat}
\vspace{-2mm}
\end{table}
\begin{table}[!t]
\caption{
Zero-shot evaluation of CLIP~\cite{radford2021learning} on retrieval and QA tasks. Results are reported on Val/Test (public) split. The best performance is highlighted in bold. 
}
\vspace{2mm}
\tablestyle{2pt}{1.1} 
\def\w{37pt} 
\scalebox{1.0}{
\begin{tabular}{c|R{\w}R{\w}R{\w}r|R{\w}R{\w}R{\w}R{\w}}
\multirow{2}{*}{ \newcellc{Input\\Channel}} & TVR & How2R & YC2R & \newcellr{VATEX-EN-R}&  TVQA   & \newcellr{How2QA} & \newcellr{VIOLIN} & VLEP  \\  
\cline{2-5}\cline{6-9} 
& AveR & AveR & AveR & AveR & Acc. & Acc.& Acc. & Acc. \\
\shline
Video-only & \bf 0.13 & \bf 0.0 & 12.61 & \bf 55.68 & \bf 27.00  & \bf 54.54 & \bf 52.40 & \bf 55.35\\
Video+Sub & \bf 0.13 & \bf 0.0 & \bf 22.61 & 46.78 & 23.17  & 44.94 & 50.00 & 56.35\\
\end{tabular}
}
\vspace{-2mm}
\label{tab:zs}
\end{table}

\subsection{Zero-Shot Evaluation on CLIP} 
Inspired by recent works~\cite{lei2021less,luo2021clip4clip} leveraging image-text pre-training for video-and-language (VidL) tasks and the strong performance of CLIP-ViT features in Section~\ref{app:vis_feat}, we further perform a zero-shot evaluation using CLIP~\cite{radford2021learning} on our retrieval and QA tasks. Captioning tasks cannot be directly evaluated as there is no decoder trained in CLIP. 

CLIP is composed of an image encoder and a text encoder. To evaluate CLIP on VidL tasks, we first sample video frames from a video clip, then encode them via the image encoder from CLIP to obtain a sequence of frame features. For subtitle and textual query, we directly apply the text encoder from  CLIP to generate a text representation. 

First, we evaluate CLIP on the video-only input. For video retrieval tasks (\emph{i.e.}, YC2R and VATEX-EN-R), we use mean pooling to aggregate the feature of all frames to obtain a global video representation. Cosine similarity is applied on the global video representation and the query representation to rank the relevance between video and query.
For video corpus moment retrieval tasks (\emph{i.e.}, TVR and How2R), an additional cosine similarity between each frame representation and query representation is used to predict the relevant span. Specifically, the localized span is determined by a sliding-window strategy. Similarly, we apply the same cosine similarity to QA tasks. For multiple-choice QA (\emph{i.e.}, TVQA and How2QA), we concatenate the question with each answer choice as query, and calculate the similarity between the global video representation and the query representation. The answer with the highest similarity score among all answer choices is selected as the predicted answer.  For VIOLIN and VLEP, which are formalized as binary classification problems, we generate the predictions according to a similarity score threshold. The best threshold is selected based on the validation set, and directly applied to the test set. 

To augment the input with subtitle channel, we simply generate the subtitle sentence representations via text encoder and max pool them to aggregate the features of all subtitle sentences to obtain a global subtitle representation. Cosine similarity is applied to global subtitle representation and query representation to obtain a similarity score. The final similarity score is defined as the unweighted average of similarities scores generated from video-only input and subtitle-only input.

Results are reported in Table~\ref{tab:zs}. Directly applying CLIP to YC2R and VATEX-EN-R achieves decent performance, which are consistent to observations in~\cite{luo2021clip4clip}. These results further support previous conclusions that image-text pre-training can benefit video-and-language tasks. However, on video moment retrieval tasks, where the model is required to localize the relevant moment based on the textual query, CLIP fails to differentiate among semantically similar video segments, resulting in poor performance. On QA tasks, where the queries or QA pairs are designed to be very similar to each other, CLIP without further finetuning struggles to predict the correct answer. Comparing video-only input to video+subtitle input, augmenting subtitle information does not guarantee performance improvement. The low performance may be due to ineffective video-subtitle fusion at prediction level or the limited capacity of CLIP to align subtitle information with textual query.

\begin{table}[!t]
\caption{
Additional results of \textbf{multi-task learning baselines} with \textbf{CLIP-ViT+SlowFast features} on Test (leaderboard) set. We compare the following model training settings: single-task training (ST), multi-task training (MT) by tasks or domains, all-task training (AT) and AT first then ST (AT $\rightarrow$ ST). The best performance (of each block) are highlighted with bold (underline).
}
\vspace{2mm}
\tablestyle{2pt}{1.1} 
\def\w{25pt} 
\resizebox{\textwidth}{!}{
\begin{tabular}{R{0.02\textwidth}l|R{\w}R{\w}R{\w}r|R{\w}R{\w}R{\w}R{\w}|R{\w}R{\w}r|R{\w}}
& \multirow{2}{*}{ \newcelll{Training\\Setting}} & TVR & How2R & YC2R & \newcellr{VATEX-\\EN-R}&  TVQA   & \newcellr{How2-\\QA} & \newcellr{VIO-\\LIN} & VLEP  & TVC & YC2C & \newcellr{VATEX-\\EN-C}& \multirow{2}{*}{\newcellr{Meta-\\Ave}}\\  
\cline{3-13}
& & AveR & AveR & AveR & AveR & Acc. & Acc.& Acc. & Acc. & C & C & C & \\
\shline
1 & Human & - & - & - & - & 89.41 & 90.32 & 91.39  & 90.50 &  62.89 & - & 62.66 & - \\
\shline
& \multicolumn{5}{l}{\textit{Finetune-only}}\\
\hline
2 & ST & 8.81 & 2.13 & 42.37 & 47.02 & 71.35  & 69.59 & 64.30 & 56.77 & \underline{50.30} & 109.89 & 55.98 & 52.59\\
3 & MT by Task  & 11.24 & 3.27 & 49.09 & 45.83 & 72.58  & 71.23 & 66.33 & 67.84 & 49.95 & 110.44 & 57.01 & 54.98\\
4 & MT by Domain  & 11.30 & 2.66 & 46.24 & 44.69 & 73.66  & 71.20 & 66.59 & 68.13 & 49.52 & 104.39 & 56.25 & 54.06\\
5 & AT & 11.98 & 3.24 & 48.40 & 46.75 & \underline{74.42}  & 71.85 & \underline{67.00} & \underline{69.06} & 49.13 & 101.76 & 56.67 & 54.57\\
6 & AT\Arrow{.2cm}ST& \underline{12.40} & \underline{3.61} &  \underline{50.93} & \bf \underline{49.91} & 74.38  & \underline{71.88} & 66.80 & 68.68 & 49.41 & \underline{110.63} & \bf\underline{58.09} & \underline{56.07}\\
\shline
& \multicolumn{5}{l}{\textit{Pretrain on ResNet+Slowfast, then Finetune}}\\
\hline
7 & ST & \bf\underline{13.70} & 3.38 & 56.59 & 46.66 & 74.52  & 73.82 & 64.19 & 67.10 & \bf \underline{51.04} & 120.22 & 55.30 & 56.96 \\
8 & MT by Task & 13.45 & \bf\underline{4.53} & \bf\underline{57.96} & 47.47 & 73.56  & 73.95 & 65.80 & 68.32 & 49.30 & 121.66 & 55.10 & 57.37\\
9 & MT by Domain  & 12.90 & 4.22 & 51.33 & 44.45 & 74.65  & 74.01 & 66.80 & 69.35 & 48.81 & 102.41 & 49.22  & 54.38\\
10 & AT & 12.55 & 3.32 & 52.16 & 46.58 & \bf \underline{75.00}  & 73.69 & \bf \underline{67.25} & 68.65 & 48.81 & 114.27 & 54.79 & 56.10\\
11 & AT\Arrow{.2cm}ST& 13.56 & 3.95 & 54.28 & \underline{49.09} & 74.83& \bf \underline{74.60}  & 67.18  & \bf\underline{69.37} & 48.13 & \bf\underline{121.89} & \underline{56.54} & \bf \underline{57.58}\\
\end{tabular}
}
\vspace{-2mm}
\label{tab:mt_test_clip_slowfast}
\end{table}
\subsection{Additional Results on Multi-Task Baselines}
Table~\ref{tab:mt_test_clip_slowfast} presents results of proposed multi-task baselines with the optimal visual representations (CLIP-ViT+SlowFast) found in Section~\ref{app:vis_feat}. The highest meta-average score of 57.58 is achieved by AT $\rightarrow$ ST with pre-training (L11). A more concise version of the table is included in VALUE leaderbaord at \url{https://value-leaderboard.github.io/leaderboard.html}.

\begin{table}[!t]
\caption{
Evaluation of \textbf{multi-task learning baselines} on Val/Test (public) split. Results are reported on HERO model with ResNet+SlowFast features unless specified otherwise. FT and PT denote finetuning and pre-training of the HERO model. We compare the following model training settings: single-task training (ST), multi-task training (MT) by tasks or domains, all-task training (AT) and AT first then ST (AT $\rightarrow$ ST). The best performance (of each block) are highlighted with bold (underline).
}
\vspace{2mm}
\tablestyle{2pt}{1.1} 
\def\w{25pt} 
\resizebox{\textwidth}{!}{
\begin{tabular}{l|R{\w}R{\w}R{\w}r|R{\w}R{\w}R{\w}R{\w}|R{\w}R{\w}r|R{\w}}
\multirow{2}{*}{ \newcellc{Training\\Setting}} & TVR & How2R & YC2R & \newcellr{VATEX-\\EN-R}&  TVQA  & \newcellr{How2-\\QA}  & \newcellr{VIO-\\LIN} & VLEP  & TVC & YC2C & \newcellr{VATEX-\\EN-C}& \multirow{2}{*}{\newcellr{Meta-\\Ave}}\\  
\cline{2-5}\cline{6-9} \cline{10-12}
& AveR & AveR & AveR & AveR & Acc. & Acc.& Acc. & Acc. & C & C & C & \\
\shline
\multicolumn{5}{l}{\textit{Finetune-only}}\\
\hline
ST & 7.72 & 1.91 & 33.91 & 58.99 & 71.08  & 69.44 & 66.83 & 58.79 & \underline{48.48} & \underline{108.46} &  \underline{52.15} & 52.52\\
MT by Task  & 7.23 & 2.70 & 39.03 & 57.64 & 71.23  & 71.65 & 66.82 & 66.64 & 47.24 & 111.35 & 51.07 & 53.87 \\
MT by Domain  & 9.91 & 3.53 & 35.76 & 73.92 & 73.89  & 71.40 & \underline{68.40} & \underline{67.51} & 47.67 & 106.44 & 50.46  & 55.35\\
AT & 9.93 & 3.29 & 38.58 & 72.84 & \underline{74.36}  & \underline{71.85} & 67.62 & 66.99 & 46.73 & 100.00 & 51.07
 & 54.96\\
AT $\rightarrow$ ST& \underline{10.53} & \underline{4.42} & \underline{41.18} & \underline{74.06} &   73.89  & 71.56 & 68.97 & 66.37 & 47.59 & 108.30 & 51.87 & \underline{56.26}\\
\shline
\multicolumn{5}{l}{\textit{Pre-train+Finetune}}\\
\hline
ST & 11.66 & \bf \underline{5.97} & 48.86 & 61.66 & 74.80 & 74.32 & 68.59 & 67.40 & \underline{50.46} & 121.89 & \underline{52.58} & 58.05 \\
MT by Task & 11.37 & 5.84 & \bf \underline{49.27} & 59.37 & 74.56  & \bf \underline{74.86} & 68.78 & 67.65 & 49.18 & \bf \underline{130.38} & 50.54 & 58.35 \\
MT by Domain  & 10.97 & 4.56 & 42.18 & 75.44 & 74.79  & 75.15 & 68.60 & 68.26 & 47.88 & 109.30 & 45.96 & 56.65\\
AT & 11.05 & 3.32 & 42.80 & 77.96 & 74.90  & 74.35 & 68.56 & \bf \underline{69.24} & 46.49 & 112.88 & 49.76 & 57.39\\
AT $\rightarrow$ ST& \underline{11.76} & 4.63 & 45.67 & \underline{78.09} & \underline{75.15}  & 74.09 & \underline{68.99} & 68.85 & 46.92 & 119.15 & 50.61 & \underline{58.54}\\
\shline
\multicolumn{5}{l}{\textit{AT $\rightarrow$ ST on CLIP-ViT+SlowFast}}\\
\hline
FT-only & 12.34 & 5.12 & 42.46 & 78.72 & 75.33  & 73.19 & 69.05 & 67.99 & \bf 50.51 & 114.60 & \bf 58.13 & 58.86\\
PT+FT & \bf 13.02 & 5.66 & 45.33 & \bf 79.95 & \bf 75.43  & 74.57 & \bf 69.40 & 69.19 & 49.67 & 115.65 & 56.35 & \bf 59.47\\
\end{tabular}
}
\vspace{-2mm}
\label{tab:multi_task_val}
\end{table}

Table~\ref{tab:multi_task_val} includes validation results of multi-task learning baselines. Table~\ref{tab:mt_full} presents more detailed results of multi-task learning baselines for retrieval and captioning tasks across different evaluation metrics on both validation split (Table~\ref{tab:full_val}) and Test (leaderboard) split (Table~\ref{tab:full_test}).

 \begin{table}[!t]
\caption{
Detailed results of multi-task training baselines on (a) Validation (Val/Test (public)) split and (b) Test (leaderboard) split of retrieval and captioning tasks. FT and PT denote finetuning and pre-training of the HERO model. We compare the following model training settings: single-task training (ST), multi-task training (MT) by tasks or domains, all-task training (AT) and AT first then ST (AT $\rightarrow$ ST). The best performance (of each block) are highlighted with bold (underline).
}
\begin{adjustbox}{angle=90}
\subfloat[Results on Validation split.
\label{tab:full_val}]{
\tablestyle{2pt}{1.2} 
\def\w{25pt} 
\scalebox{.82}{
\begin{tabular}{l|R{\w}R{\w}R{\w}|R{\w}R{\w}R{\w}|R{\w}R{\w}R{\w}|R{\w}R{\w}R{\w}|R{\w}R{\w}R{\w}R{\w}|R{\w}R{\w}R{\w}R{\w}|R{\w}R{\w}R{\w}R{\w}}
\multirow{2}{*}{ \newcellc{Training\\Setting}} & \multicolumn{3}{c|}{TVR
}& \multicolumn{3}{c|}{How2R} & \multicolumn{3}{c|}{YC2R} & \multicolumn{3}{c|}{VATEX-EN-R}&  \multicolumn{4}{c|}{TVC} & \multicolumn{4}{c|}{YC2C} & \multicolumn{4}{c}{VATEX-EN-C}\\  
\cline{2-25}
& R@1 & R@5 & R@10  & R@1 & R@5 & R@10  &  R@1 & R@5 & R@10  & R@1 & R@5 & R@10 & B@4 & R & M & C & B@4 & R & M & C & B@4 & R & M & C\\
\shline
Human & - & - & - & - & - & - & - & - & - & - & - & - & 12.90 & 36.50 & 20.60 & 64.56 & - & - & - & - & 24.65 & 46.85 & 24.89 & 61.86\\
\shline
\multicolumn{5}{l}{\textit{Finetune-only}}\\
\hline
ST & 3.38 & 8.63 & 11.14  & 0.62 & 2.01 & 3.09  & 18.73 & 37.17 & 45.82 & 29.72 & 67.48 & 79.77  & \underline{11.31} & \underline{33.31} & \underline{17.21} & \underline{48.48} & 9.64 & \underline{37.15} & 16.98 & 108.46 & 29.64 & 48.03 & \underline{22.70} & \underline{52.15} \\
MT by Task  & 2.47 & 7.88 & 11.34  & 0.77 & 3.09 & 4.25  & 22.51 & 42.73 & 51.86 & 29.28 & 65.92 & 77.75 & 10.72 & 32.76 & 17.07 & 47.24 & \underline{9.87} & 36.71 & 17.03 & \underline{111.35} & 28.87 & 47.33 & 22.50 & 51.07\\
MT by Domain  &  4.11 & 10.82 & 14.79  & 1.70 & 3.86 & 5.02  & 18.19 & 39.64 & 49.46 & 45.50 & \underline{84.17} & 92.08 & 10.99 & 33.20 & 17.14 & 47.67 & 9.45 & 36.45 & 16.53 & 106.44 & 29.67 & 47.75 & 22.32 & 50.46\\
AT & 4.23 & 10.77 & 14.78  & 1.00 & 3.48 & 5.40  & 20.90 & 43.01 & 53.06 & 44.12 & 83.00 & 91.80 & 10.43 & 32.61 & 16.88 & 46.96 & 9.28 & 35.82 & 16.45 & 102.26 & \underline{30.14} & 48.11 & 22.45 & 51.00\\
AT $\rightarrow$ ST& \underline{4.67} & \underline{11.46} & \underline{15.45}  & \underline{1.78} & \underline{5.17} & \underline{6.64}  & \underline{23.17} & \underline{45.31} & \underline{55.07} & \underline{45.66} & 84.10 & \underline{92.42} & 10.50 & 32.85 & 16.96 & 47.59 & 9.59 & 37.05 & \underline{17.06} & 108.30 & 29.74 & \underline{48.18} & 22.60 & 51.87\\
\shline
\multicolumn{5}{l}{\textit{Pre-train+Finetune}}\\
\hline
ST &  \underline{5.57} & 12.43 & 16.99  & \bf \underline{3.01} & \bf \underline{6.33} & \bf \underline{8.57}  & \bf \underline{31.30} & 53.04 & 62.23 & 33.04 & 70.31 & 81.64 & \bf\underline{12.25} & \bf \underline{34.10} & \underline{17.54} & \underline{50.46} & 11.47 & 39.79 & 18.14 & 121.89 & \underline{30.07} & \underline{48.34} & \underline{22.72} & \underline{52.58}\\
MT by Task & 5.17 & 12.16 & 16.79  & 2.78 & 6.18 & \underline{8.57}  & 30.18 & \bf \underline{54.13} & \bf \underline{63.49} & 31.17 & 67.62 & 79.31 & 11.53 & 33.61 & 17.42 & 49.18 & \bf \underline{12.40} & \bf \underline{40.41} & \bf \underline{18.81} & \bf \underline{130.38} & 29.22 & 47.57 & 22.37 & 50.54\\
MT by Domain  &  4.61 & 11.82 & 16.48  & 2.78 & 4.63 & 6.27  & 22.34 & 46.65 & 57.56 & 48.07 & 85.27 & 92.99 & 11.79 & 33.75 & 17.15 & 47.88 & 10.09 & 38.27 & 17.40 & 109.30 & 28.98 & 48.01 & 21.86 & 45.96\\
AT & 5.08 & 12.08 & 15.98  & 1.47 & 3.55 & 4.94  & 23.80 & 47.31 & 57.30 & 51.72 & 87.90 & \underline{94.26} & 10.92 & 32.96 & 16.86 & 46.49 & 10.54 & 38.04 & 17.36 & 112.88 & 29.72 & 48.13 & 22.26 & 49.76\\
AT $\rightarrow$ ST& 5.49 & \underline{12.61} & \underline{17.18}  & 2.62 & 5.02 & 6.25  & 27.38 & 50.09 & 59.53 & \underline{52.08} & \underline{87.92} & \underline{94.26} & 11.01 & 33.06 & 16.94 & 46.92 & 11.17 & 38.89 & 17.82 & 119.15 & 29.66 & 48.19 & 22.52 & 50.61\\
\shline
\multicolumn{5}{l}{\textit{AT $\rightarrow$ ST on ViT+SlowFast}}\\
\hline
FT-only &  5.67 & 13.44 & 17.91  & 2.78 & 5.33 & 7.26  & 24.71 & 46.31 & 56.36 & 53.70 & 88.20 & 94.26 & \underline{11.68} & \underline{33.91} & \bf \underline{17.63} & \bf \underline{50.51} & 10.35 & 37.93 & 17.46 & 114.60 & \bf \underline{32.93} & \bf \underline{50.13} & \bf \underline{24.06} & \bf \underline{58.13}\\
PT+FT &  \bf\underline{5.93} & \bf \underline{14.36} & \bf \underline{18.76}  & \bf \underline{3.01} & \underline{6.18} & \underline{7.80} & \underline{26.23} & \underline{49.71} & \underline{60.05} & \bf \underline{55.62} & \bf \underline{89.15} & \bf \underline{95.07} & 11.52 & 33.86 & 17.44 & 49.67 & \underline{10.62} & \underline{38.59} & \underline{17.70} & \underline{115.65} & 32.69 & 50.02 & 23.85 & 56.35\\
\end{tabular}
}
}
\end{adjustbox}
\begin{adjustbox}{angle=90}
\subfloat[Results on Test (leaderboard) split.
\label{tab:full_test}]{
\tablestyle{2pt}{1.2} 
\def\w{25pt} 
\scalebox{.82}{
\begin{tabular}{l|R{\w}R{\w}R{\w}|R{\w}R{\w}R{\w}|R{\w}R{\w}R{\w}|R{\w}R{\w}R{\w}|R{\w}R{\w}R{\w}R{\w}|R{\w}R{\w}R{\w}R{\w}|R{\w}R{\w}R{\w}R{\w}}
\multirow{2}{*}{ \newcellc{Training\\Setting}} & \multicolumn{3}{c|}{TVR
}& \multicolumn{3}{c|}{How2R} & \multicolumn{3}{c|}{YC2R} & \multicolumn{3}{c|}{VATEX-EN-R}&  \multicolumn{4}{c|}{TVC} & \multicolumn{4}{c|}{YC2C} & \multicolumn{4}{c}{VATEX-EN-C}\\  
\cline{2-25}
& R@1 & R@5 & R@10  & R@1 & R@5 & R@10  &  R@1 & R@5 & R@10  & R@1 & R@5 & R@10 & B@4 & R & M & C & B@4 & R & M & C & B@4 & R & M & C\\
\shline
Human & - & - & - & - & - & - & - & - & - & - & - & - & 13.05 & 36.52 & 20.65 & 62.89 & - & - & - & - & 24.77 & 46.88 & 25.09 & 62.66\\
\shline
\multicolumn{5}{l}{\textit{Finetune-only}}\\
\hline
ST & 3.10 & 8.44 & 11.55 & 0.32 & 1.74 & 3.16 & 24.00 & 44.26 & 53.80 & 15.97 & 42.14 & \underline{56.91} & \underline{11.26} & \underline{32.96} & 16.91 & \underline{46.76} & 9.00 & \underline{36.35} & 16.58 & \underline{106.24} & 29.63 & 48.02 & \underline{22.70} & \underline{52.16} \\
MT by Task  & 2.99 & 8.21 & 12.06  & 0.24 & 1.90 & 3.56  & 27.62 & 51.49 & 60.04 & 16.45 & 41.99 & 56.08 & 10.69 & 32.60 & 16.87 & 46.01 & \underline{9.54} & 35.51 & \underline{16.62} & 105.22 & 28.86 & 47.32 & 22.49 & 51.07\\
MT by Domain  & 4.35 & 10.79 & 14.88  & 1.18 & \underline{3.01} & 3.87  & 24.13 & 50.25 & 59.35 & 14.80 & 39.53 & 53.96 &  11.05 & \underline{32.96} & \underline{16.96} & 46.53 & 8.61 & 35.20 & 16.24 & 100.74 & 29.67 & 47.74 & 22.31 & 50.46 \\
AT & 4.30 & 10.57 & 14.42   & 0.71 & 2.45 & \underline{4.11} & 27.81 & 53.57 & 62.34 & 15.63 & 40.80 & 55.56 & 10.89 & 32.75 & 16.84 & 46.46 & 9.18 & 35.21 & 16.27 & 101.72 & \underline{30.21} & 48.10 & 22.49 & 51.07\\
AT $\rightarrow$ ST& \underline{4.82} & \underline{11.43} & \underline{15.03}  & \underline{1.34} & 2.84 & 4.03  & \underline{31.17} & \underline{54.11} & \underline{63.15} & \underline{16.52} & \underline{42.42} & 56.79 & 10.65 & 32.69 & 16.75 & 46.12 & 9.00 & 36.32 & 16.67 & 103.73 & 29.74 & \underline{48.20} & 22.61 & 51.87\\
\shline
\multicolumn{5}{l}{\textit{Pre-train+Finetune}}\\
\hline
ST & 5.69 & 13.38 & 17.06  & 1.90 & 4.43 & 5.93  & 38.97 & 63.65 & 71.01 & \bf\underline{18.33} & \bf\underline{44.77} & \bf\underline{58.78} & \bf\underline{12.00} & \bf\underline{33.94} & \underline{17.39} & \underline{48.97} & 11.46 & \bf\underline{39.67} & \bf\underline{18.17} & \bf\underline{127.94} & \underline{30.07} & \underline{48.32} & \underline{22.80} & \underline{52.57}  \\
MT by Task & \underline{6.01} & \underline{13.71} & \underline{18.18} & 1.82 & \bf \underline{5.06} & \bf \underline{7.11}  & \bf\underline{39.46} & \bf\underline{63.90} & \bf\underline{74.25} & 17.95 & 43.84 & 58.12 & 11.56 & 33.42 & 17.21 & 48.02 & \bf\underline{11.70} & 38.90 & 18.13 & 123.40 & 29.21 & 47.58 & 22.36 & 50.49  \\
MT by Domain  & 5.37 & 12.46 & 16.76 & 2.21 & 3.95 & 5.93 & 30.42 & 57.54 & 68.45 & 15.73 & 40.60 & 54.58  & 11.93 & 33.61 & 17.04 & 47.23 & 9.11 & 36.60 & 16.87 & 100.29 & 28.99 & 47.98 & 21.85 & 45.95 \\
AT & 5.35 & 12.61 & 16.87 & 2.14 & 4.03 & 5.93 & 31.86 & 57.54 & 67.21 & 16.48 & 41.40 & 56.15 & 11.19 & 33.10 & 16.84 & 46.04 & 10.17 & 37.40 & 17.21 & 109.11 & 29.70 & 48.13 & 22.26 & 49.74\\
AT $\rightarrow$ ST& 5.72 & 13.21 & 17.57  & \bf \underline{2.29} & 4.98 & 6.25  & 34.73 & 59.85 & 67.89 & 17.08 & 42.59 & 56.90 & 11.40 & 33.09 & 16.88 & 46.38 & 11.32 & 39.03 & 17.83 & 120.86 & 29.62 & 48.18 & 22.52 & 50.59\\
\shline
\multicolumn{5}{l}{\textit{AT $\rightarrow$ ST on CLIP-ViT+SlowFast}}\\
\hline
FT-only & 5.91 & 13.58 & 17.72  & 1.42 & \underline{3.79} & 5.61  & 31.73 & 55.67 & 65.40 & \underline{25.01} & \underline{55.36} & \underline{69.36} & 11.61 & \underline{33.84} & \bf\underline{17.49} & \bf\underline{49.41} & 9.94 & 37.62 & 17.59 & 110.63 & \bf\underline{32.89} & \bf\underline{50.01} & \bf\underline{24.04} & \bf\underline{58.09} \\
PT+FT & \bf \underline{6.39} & \bf \underline{14.75} & \bf \underline{19.54} & \underline{1.90} & \underline{3.79} & \underline{6.17}  & \underline{35.10} & \underline{59.48} & \underline{68.27} & 24.51 & 54.34 & 68.41  & \underline{11.78} & 33.68 & 17.15 & 48.13 & \underline{11.25} & \underline{39.02} & \underline{17.92} & \underline{121.89} & 32.68 & \bf\underline{50.01} & 23.84 & 56.29 \\
\end{tabular}
}
}
\end{adjustbox}
\vspace{-2mm}
\label{tab:mt_full}
\end{table}

\section{Collection of Human Baselines}\label{sec:human_eval}
\begin{figure}[b!]
    \centering
    \includegraphics[width=.9\textwidth]{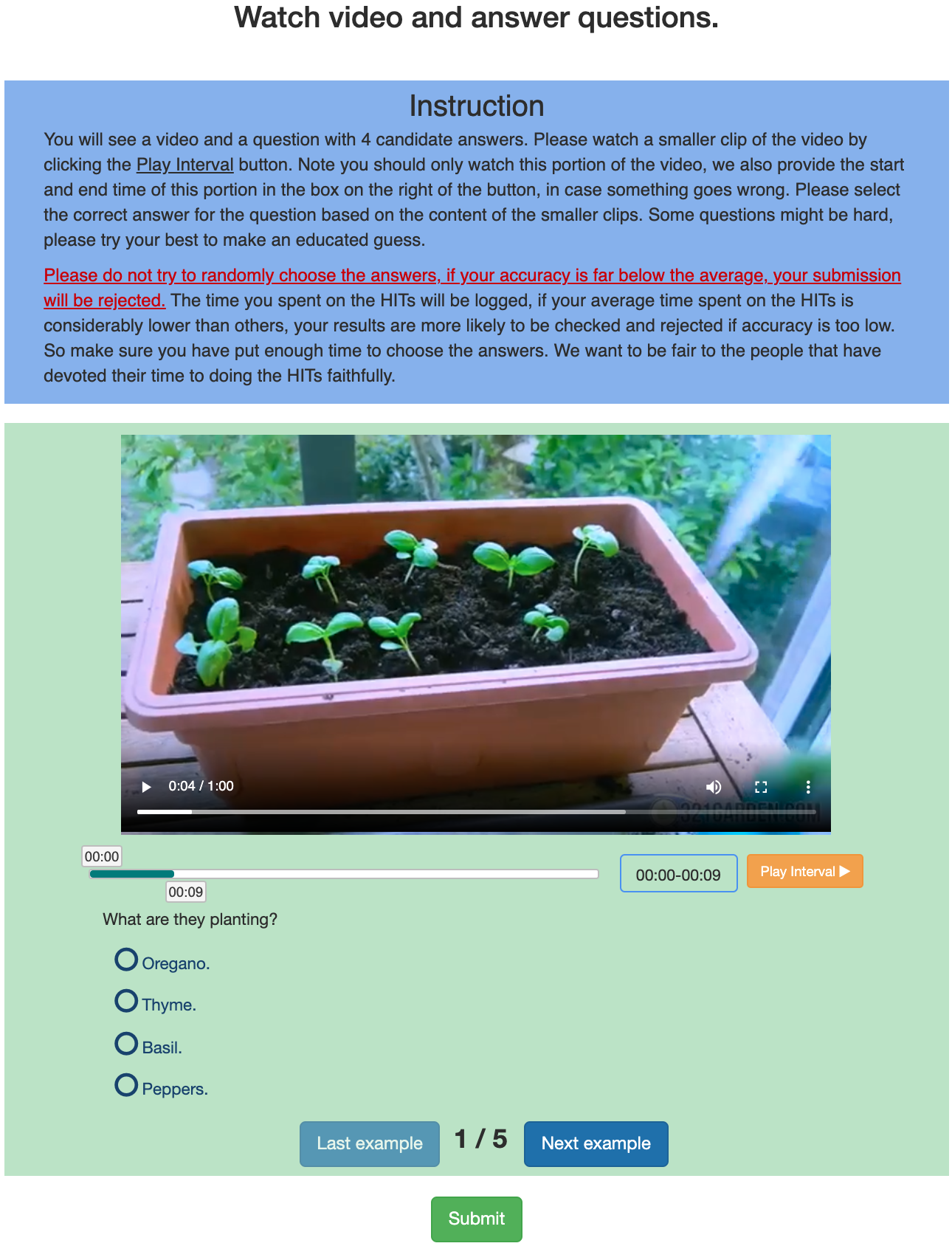}
    \caption{UI for huamn evaluation on video QA task.}
    \label{fig:human_eval}
\end{figure}

For multiple-choice QA tasks (\emph{i.e.}, TVQA and How2QA), we resort to crowd-sourcing to obtain human baselines. Specifically, we present the human annotator with a multi-channel video, a question about the video, and a set of answer candidates. The annotator is asked to select the correct answer. Each question is presented to one annotator to evaluate human performance. For VIOLIN, a pair of video and hypothesis is presented to 3 human annotators, who are asked to determine whether the hypothesis is entailed or contradict to the video content. The human performance is evaluated based on the majority vote across the 3 human responses. For VLEP, human annotators are required to choose a more likely event from a pair of next event candidates based on the video content. We also take the majority vote to evaluate human performance. An example of our human evaluation interface is shown in Figure~\ref{fig:human_eval}. The estimated hourly pay to our annotators is  \$8.6. The total amount spent is \$2173.4.

For captioning tasks, we randomly sample one caption from the ground-truth annotations and use the rest as references to calculate the human performance across all captioning metrics. Note that in YC2C, there is only one caption collected for each video clip, thereby human performance is not reported. 

\section{Additional Experimental Details}\label{app:model}
\subsection{Downstream Adaptation} 
We describe in detail how HERO~\cite{li2020hero} architecture can be adapted to VALUE tasks.

For retrieval tasks, we add a query encoder head, consisting of a self-attention layer, two linear layers and an LN layer, on top of HERO's cross-modal transformer to obtain the query embeddings. The input multi-channal videos are encoded by cross-modal transformer and temporal transformer in HERO to obtain the contextualized video embeddings. 
For video moment retrieval tasks (TVR~\cite{lei2020tvr} and How2R~\cite{li2020hero}), we follow XML~\cite{lei2020tvr} to compute the matching scores between the query and visual frames both locally (frame-level, for moment retrieval) and globally (clip-level, for video retrieval). Specifically, we use cross-entropy loss to supervise the learning of the start and end index for local alignment and a combined hinge loss~\cite{yu2018mattnet} over positive and negative query-video pairs for global alignment. For video retrieval tasks (YC2R~\cite{zhou2018towards} and VATEX-EN-R~\cite{wang2019vatex}), only the combined hinge loss is adopted. 

For multiple-choice QA tasks (TVQA~\cite{lei2018tvqa} and How2QA~\cite{li2020hero}), we append the corresponding QA pair (question and an answer candidate) to each of the subtitle sentences, which is fed into the cross-modal transformer to perform early fusion with local textual context. In addition, these QA pairs are also appended to the input of temporal transformer to be fused with global video context. We use a simple attention layer to compute the weighted-sum-across-time of the QA-aware frame representations from the temporal transformer output. These final QA-aware global representations are then fed through an MLP and softmax layer to obtain the probability score  of all the answers for the corresponding question. cross-entropy loss is used to supervise the model training. When supervision is available,\footnote{For example, TVQA and How2QA provides start and end timestamps to localize `frames of interest' for the question.} a span prediction loss (addition of two cross-entropy loss on start and end timestamps) is added as additional supervision. 

Similar to multiple-choice QA, we append each natural language hypothesis in VIOLIN~\cite{liu2020violin} (or next event candidate in VLEP~\cite{lei2020more}) to each of the subtitle sentences, as well as to the input of Temporal Transformer. A simple attention pooling layer is added to HERO to obtain the final query-aware global representations. We apply cross-entropy loss for the training.

For captioning tasks, a Transformer decoder~\cite{vaswani2017attention} is employed to empower HERO with generative capabilities.
We feed the whole subtitle-aligned video clip into HERO and obtain the subtitle-fused video representation for each frame. 
For TVC~\cite{lei2020tvr}, frame representations are further grouped by the ``moment of interest" using the time interval provided in the caption annotation, and the decoder-to-encoder attention is applied on the representations of the corresponding video moment. For YC2C~\cite{zhou2018towards} and VATEX-EN-C~\cite{wang2019vatex}, as the caption is to describe the whole clip, the decoder-to-encoder attention is applied on the representations of the entire video. 
The decoder is trained with conventional left-to-right language modeling cross-entropy loss together with the HERO encoder end-to-end.
We follow MMT~\cite{lei2020tvr} to use shallow Transformer decoder (2-layer) with 768-D hidden size, and simply use the greedy decoding at inference for constructing the baselines. 

\subsection{Video-Subtitle Fusion Methods}
We introduce three early fusion baselines in detail. Let's denote the video segments embeddings as $\mathbf{F}_{\mathbf{V}} = \{f_{v}, v \in \mathbf{V}\}$ and subtitle sentence embeddings as $\mathbf{F}_{\mathbf{S}} = \{f_{s}, s \in \mathbf{S}\}$. 
The video segments embeddings are the concatenations of pre-extracted 2D appearance features concatenated with 3D motion features for each video segment. The subtitle sentence embeddings are obtained by max-pooling the contextualized subtitle token embeddings from a multi-layer transformer for each subtitle sentence. 
The first method (\textit{sequence concat}) concatenates embeddings at sequence level without temporal alignment, denoted as $\mathbf{F}_{\mathbf{V}}|\mathbf{F}_{\mathbf{S}}$. 
The second method (\textit{temporal align + sum}) takes the summation of the temporally aligned visual frame embeddings and subtitle sentence embeddings, denoted as $f_{v}+f_{s}$. 
The third method (\textit{temporal align + concat}) concatenates the temporally aligned visual frame embeddings with subtitle sentence embeddings at feature level, denoted as $f_{v}|f_{s}$. Compared to HERO, we simply replace cross-modal transformer with methods described above. The fused embeddings from all the methods above are then fed into the temporal transformer to learn the global video context and obtain the final video embeddings. 

\subsection{Multi-Task Baselines} 
All our multi-task models are trained with a shared HERO encoder. We add only one head for each task type. For example, the same Transformer decoder is shared among different captioning tasks.

\subsection{Implementation Details}

Our models are implemented based on PyTorch~\cite{paszke2017automatic}.\footnote{\url{https://pytorch.org/}} To speed up training, we use Nvidia Apex\footnote{\url{https://github.com/NVIDIA/apex}} for mixed precision training. Gradient accumulation~\cite{ott2018scaling} is applied to reduce multi-GPU communication overheads. All experiments are run on 4 or 8 Nvidia V100 GPUs (32GB VRAM; NVLink connection) on Microsoft Azure.\footnote{\url{https://azure.microsoft.com/}} 
We use AadmW~\cite{AdamW} to optimize model parameters, with an initial learning rate in $\{3e-5, 5e-5, 1e-4\}$, $\beta_1{=}0.9$, $\beta_2{=}0.98$, and use learning rate warmup over the first 10\% training steps followed by linear decay to 0. 

For single-task training, since the considered datasets vary in scale and domain, we use task-specific learning rates and training steps based on validation performance for each dataset. For multi-task training, we sample one task per mini-batch to train with a probability approximately proportional to the number of training examples for each task. The best checkpoint is selected based on the highest meta-average score achieved on validation split. To reproduce our results, please check the released starter code at  \url{https://github.com/VALUE-Leaderboard/StarterCode}.

For YC2 and VATEX datasets, we employ ASR tool from Azure Cognitive Service\footnote{\url{https://azure.microsoft.com/en-us/services/cognitive-services/speech-services/}} to generate the subtitles.

\section{License and Usage}\label{sec:content_license_usage}
As per the original authors, the annotations for TVQA~\cite{lei2018tvqa}, TVR~\cite{lei2020tvr}, TVC~\cite{lei2020tvr}, VIOLIN~\cite{liu2020violin}, YouCookII~\cite{zhou2018towards}, VLEP~\cite{lei2020more}, How2QA~\cite{li2020hero}, How2R~\cite{li2020hero} are under CC BY-NC-SA 4.0 license\footnote{\url{https://creativecommons.org/licenses/by-nc-sa/4.0/}}, the annotations for VATEX~\cite{wang2019vatex} are under CC BY 4.0~\footnote{\url{https://creativecommons.org/licenses/by/4.0/}}.
The videos used in the datasets are from TV shows and YouTube, on non-offensive topics such as sitcoms and instructional videos.
The annotations in the datasets do not contain personally identifiable information.
Our released features are under CC BY-NC-SA 4.0 license\footnote{\url{https://creativecommons.org/licenses/by-nc-sa/4.0/}}, and our code is under MIT license\footnote{\url{https://opensource.org/licenses/MIT}}.  

The datasets used in the benchmark contain biases, both in the videos and the annotations. Such biases might be reflected in the predictions of the systems trained on these data. Users should not completely rely on such systems for making real-world decisions.

\end{document}